\documentclass[letterpaper, 10 pt, conference]{ieeeconf}  

\IEEEoverridecommandlockouts                              

\usepackage{times}
\usepackage{epsfig}
\usepackage{graphicx}
\usepackage{float}
\usepackage{wrapfig}

\usepackage{amsmath,amssymb}

\usepackage{bm,xspace}
\usepackage{comment}
\usepackage{verbatim}
\usepackage{multirow}
\usepackage{multicol}
\usepackage{balance}
\usepackage{url}
\usepackage{booktabs}
\usepackage{etoolbox,siunitx}
\usepackage{calc}
\usepackage{pifont,hologo}
\usepackage{color}

\usepackage[super]{nth}
\usepackage{nicefrac}
\def\tt{\mathbf{t}}

\newcommand{\YesV}{\ding{51}}%
\newcommand{\NoX}{\ding{55}}%

\DeclareMathSymbol{@}{\mathord}{letters}{"3B}

\newcommand\mypara[1]{\vspace{3pt}\noindent\textbf{#1.}}

\def\rot#1{\rotatebox{90}{#1}}

\def\latex/{\LaTeX}
\def\bibtex/{\hologo{BibTeX}}

\usepackage[official]{eurosym}

\newcommand{\ie}{\emph{i.e.~}}
\newcommand{\eg}{\emph{e.g.~}}
\newcommand{\etc}{\emph{etc.~}}
\newcommand{\etal}{\emph{et al.}}

\newcommand{\figLabel}{Figure\xspace}

\newcommand{\secLabel}{Section\xspace}
\newcommand{\tblLabel}{Table\xspace}

\usepackage{hyperref}
\hypersetup{
    colorlinks=true,
    linkcolor=magenta,
    filecolor=magenta,      
    urlcolor=blue,
    pdftitle={OpenBot},
    pdfpagemode=FullScreen,
    }
    
\graphicspath{{./figures/}}

\title{\LARGE \bf
OpenBot: Turning Smartphones into Robots
}

\author{Matthias M\"uller$^{1}$ 
and Vladlen Koltun$^{1}$

\thanks{$^{1}$Intelligent Systems Lab, Intel}
}

\begin{document}

\maketitle
\thispagestyle{empty}
\pagestyle{empty}

\begin{abstract}
Current robots are either expensive or make significant compromises on sensory richness, computational power, and communication capabilities. We propose to leverage smartphones to equip robots with extensive sensor suites, powerful computational abilities, state-of-the-art communication channels, and access to a thriving software ecosystem. We design a small electric vehicle that costs \$50 and serves as a robot body for standard Android smartphones. We develop a software stack that allows smartphones to use this body for mobile operation and demonstrate that the system is sufficiently powerful to support advanced robotics workloads such as person following and real-time autonomous navigation in unstructured environments. Controlled experiments demonstrate that the presented approach is robust across different smartphones and robot bodies. 
\end{abstract}


\section{Introduction}
Robots are expensive.
Legged robots and industrial manipulators cost as much as luxury cars, and wheeled mobile robots such as the popular Husky A200 still cost \$20K. Even the cheapest robots from Boston Dynamics, Franka Emika or Clearpath cost at least \$10K. Few academic labs can afford to experiment with robotics on the scale of tens or hundreds of robots.

A number of recent efforts have proposed designs for more affordable robots.
Kau \etal~\cite{kau2019stanford} and Grimminger \etal~\cite{grimminger2019open} proposed quadruped robots that rely on low-cost actuators and cost \$3K and \euro{4K}. Yang \etal~\cite{yang2019replab}, Gupta \etal~\cite{gupta2018robot}, and Gealy \etal~\cite{gealy2019quasi} proposed manipulation robots that cost \$2K, \$3K, and \$5K respectively.
A number of mobile robots for hobbyist and researchers have been released which fall in the \$250--500 range. These include the AWS DeepRacer~\cite{balaji2019deepracer}, the DJI Robomaster S1~\cite{robomaster_s1}, the Nvidia JetBot~\cite{nvidia_jetbot}, and the DukieBot~\cite{paull2017duckietown}.
In order to achieve this price point, these platforms had to make compromises with regards to the physical body, sensing, communication, and compute. 
Is there an alternative where robots become extremely cheap, accessible to everyone, and yet possess extensive sensory abilities and computational power?

In this work, we push further along the path to highly capable mobile robots that could be deployed at scale. Our key idea is to leverage smartphones.
We are inspired in part by projects such as Google Cardboard: by plugging standard smartphones into cheap physical enclosures, these designs enabled millions of people to experience virtual reality for the first time. Can smartphones play a similar role in robotics?

More than 40\% of the world's population own smartphones.
Commodity models now carry HD cameras, powerful CPUs and GPUs, advanced IMUs, GPS, WiFi, Bluetooth, 4G modems, and more. Modern smartphones are even equipped with dedicated AI chips for neural network inference, some of which already outperform common desktop processors~\cite{ignatov2019ai}.

\begin{figure}[!tb]
\centering
 \includegraphics[trim=50 800 0 1000, clip, width=\linewidth]{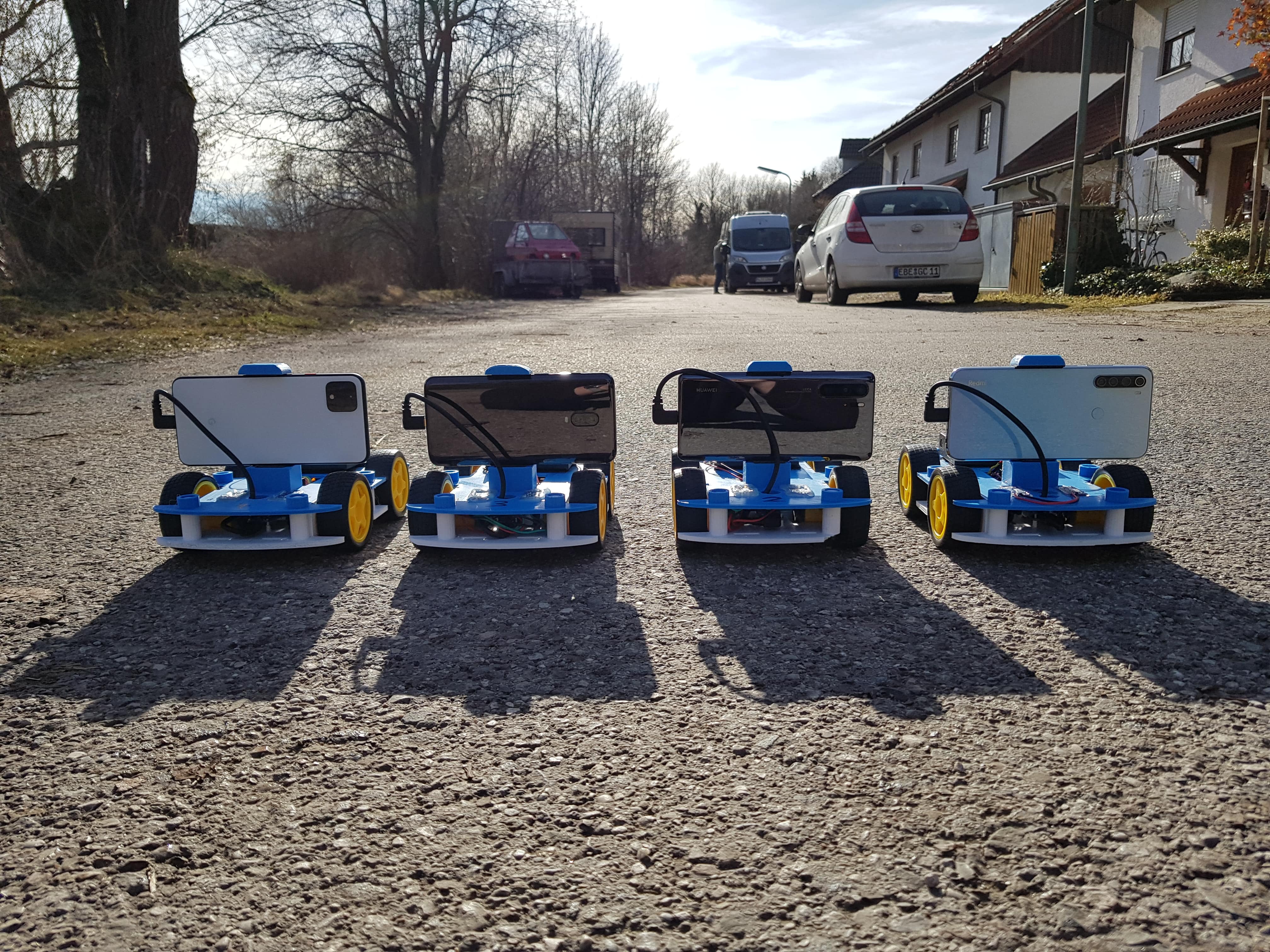} \\
 \caption{\textbf{OpenBots.} Our wheeled robots leverage a smartphone for sensing and computation. The robot body costs \$50 without the smartphone. The platform supports person following and real-time autonomous navigation in unstructured environments.}
\vspace{-8pt}
\label{fig:intro}
\end{figure}

We develop and validate a design for a mobile robot that leverages a commodity smartphone for sensing and computation (\figLabel \ref{fig:intro}). The smartphone acts as the robot's brain and sensory system. This brain is plugged into a cheap electromechanical body that costs less than \$50. 

Using off-the-shelf smartphones as robot brains has numerous advantages beyond cost.
Hardware components on custom robots are quickly outdated.
In contrast, consumer-grade smartphones undergo generational renewal on an annual cadence, acquiring cameras with higher resolution and higher frame rate, faster processors, new sensors, and new communication interfaces.
As a side effect, second-hand smartphones are sold cheaply, ready for a second life as a robot.
In addition to the rapid advancement of hardware capabilities, smartphones benefit from a vibrant software ecosystem. 
Our work augments this highly capable bundle of sensing and computation with a mobile physical body and a software stack that supports robotics workloads. 

Our work makes four contributions.
(1) We design a small electric vehicle that relies on cheap and readily available components with a hardware cost of only \$50 as a basis for a low-cost wheeled robot. (2) We develop a software stack that allows smartphones to use this vehicle as a body, enabling mobile navigation with real-time onboard sensing and computation.
(3) We show that the proposed system is sufficiently powerful to support advanced robotics workloads such as person following and autonomous navigation.
(4) We perform extensive experiments that indicate that the presented approach is robust to variability across smartphones and robot bodies.

Our complete design and implementation, including all hardware blueprints and the software suite are freely available at \url{https://www.openbot.org} to support affordable robotics research and education at scale.

\section{Related Work}
Wheeled robots used for research can be divided into three main categories: tiny robots used for swarm robotics, larger robots based on RC trucks used for tasks that require extensive computation and sensing, and educational robots.
Swarm robots~\cite{rubenstein2012kilobot,mclurkin2014robot,wilson2016pheeno} are inexpensive but have very limited sensing and compute. They are designed to operate in constrained indoor environments with emphasis on distributed control and swarm behaviour. On the other end of the spectrum are custom robots based on RC trucks~\cite{gonzales2016autonomous, codevilla2018end, srinivasa2019mushr, o2019f1, balaji2019deepracer, goldfain2019autorally}. They feature an abundance of sensors and computation, supporting research on problems such as autonomous navigation and mapping. However, they are expensive and much more difficult to assemble and operate. Educational robots \cite{riedo2013thymio, paull2017duckietown} are designed to be simple to build and operate while maintaining sufficient sensing and computation to showcase some robotic applications such as lane following. However, their sensors and compute are usually not sufficient for cutting-edge research. Some robots such as the DuckieBot~\cite{paull2017duckietown} and Jetbot~\cite{nvidia_jetbot} try to bridge this gap with designs that cost roughly \$250. However, these vehicles are small and slow. In contrast, our wheeled robot body costs \$50 or less and has a much more powerful battery, bigger chassis, and four rather than two motors. The body serves as a plug-in carrier for a smartphone, which provides computation, sensing, and communication. Leveraging off-the-shelf smartphones allows this design to exceed the capabilities of much more expensive robots.

Contemporary smartphones are equipped with mobile AI accelerators, the capabilities of which are rapidly advancing. Ignatov \etal~\cite{ignatov2019ai} benchmark smartphones with state-of-the-art neural networks for image classification, image segmentation, object recognition, and other demanding workloads. Not only are most recent smartphones able to run these complex AI models, but they approach the performance of CUDA-compatible graphics cards. Lee \etal~\cite{lee2019device} show how to leverage mobile GPUs that are already available on most smartphones in order to run complex AI models in real time. They also discuss design considerations for optimizing neural networks for deployment on smartphones. Our work harnesses these consumer hardware trends for robotics. 

There have been a number of efforts to combine smartphones and robotics. In several hobby projects, smartphones are used as a remote control for a robot \cite{phonebot15, androidRC16}. On Kickstarter, Botiful~\cite{botiful12} and Romo~\cite{romo12} raised funding for wheeled robots with smartphones attached for telepresence, and Ethos~\cite{ethos15} for a drone powered by a smartphone. Most related to our work is Wheelphone~\cite{wheelphone13}, where a smartphone is mounted on a robot for autonomous navigation. Unfortunately, this project is stale; the associated Github repos have only 1 and 4 stars and the latest contribution was several years ago. The robot has two motors providing a maximum speed of only 30 cm/s and is restricted to simple tasks such as following a black tape on the floor or obstacle avoidance on a tabletop. Despite these drawbacks, it costs \$250. Our robot is more rugged, can reach a maximum speed of 150 cm/s, costs \$30-50, and is capable of heavy robotic workloads such as autonomous navigation.

Researchers have also explored the intersection of smartphones and robotics.  Yim \etal~\cite{yim2010} detect facial expressions and body gestures using a smartphone mounted on a robot to study social interaction for remote communication via a robotic user interface. 
DragonBot \cite{setapen2012creating} is a cloud-connected 5-DoF toy robot for studying human/robot interaction; a smartphone is used for control and a visual interface.
V.Ra~\cite{cao2019v} is a visual and spatial programming system for robots in the IoT context. Humans can specify a desired trajectory via an AR-SLAM device (\eg a smartphone) which is then attached to a robot to execute this trajectory. In contrast to our work, the navigation is not autonomous but relies on user input. 
Oros \etal~\cite{oros2013smartphone} leverage a smartphone as a sensor suite for a wheeled robot. The authors retrofit an RC truck with a smartphone mount and I/O interface to enable autonomous operation. However, the key difference is that they stream the data back to a computer for processing. Moreover, the proposed robot costs \$350 without the smartphone and does not leverage recent advancements that enable onboard deep neural network inference or visual-inertial state estimation. The project is stale, with no updates to the software in 7 years. 

In summary, the aforementioned projects use the smartphone as a remote control for teleoperation, offload data to a server for processing, or rely on commercial or outdated hardware and software. In contrast, our platform turns a smartphone into the brain of a fully autonomous robot with onboard sensing and computation.

\section{System}

\begin{table}[!b]
\setlength{\tabcolsep}{9pt}
\centering
\begin{tabular}{lccc}
\toprule
\textbf{Component} & \textbf{Quantity} & \textbf{Unit Price} & \textbf{Bulk Price}\\ 
\midrule
3D-printed Body & 1 & \$5.00 & \$5.00 \\
Arduino Nano & 1 & \$8.00 & \$3.00\\ 
Motor + Tire & 4 & \$3.50 & \$2.00 \\ 
Motor Driver & 1 & \$3.00 & \$2.00\\ 
Battery (18650) & 3 & \$5.00 & \$3.00\\
Speed Sensor & 2 & \$1.50 & \$1.00\\
Sonar Sensor & 1 & \$2.00 & \$1.00\\
Miscellaneous* & 1 & \$5.00 & \$3.00 \\
\midrule
\textbf{Total} & & \textbf{\$50} & \textbf{\$30} \\
\bottomrule
\end{tabular}
\caption{\textbf{Bill of materials.} Unit price is the approximate price per item for a single vehicle. The bulk price is the approximate price per item for five vehicles. *Not included in total cost (screws, wires, resistors, LEDs, spring, etc.)
}
\vspace{-8pt}
\label{tbl:bom}
\end{table}

\begin{figure*}[!htb]
\centering
\vspace{4pt}
\begin{tabular}{@{}c@{\hspace{5mm}}c@{\hspace{3mm}}c@{\hspace{3mm}}c@{}}
Top & Front & Side & Back \\
\includegraphics[height=3cm]{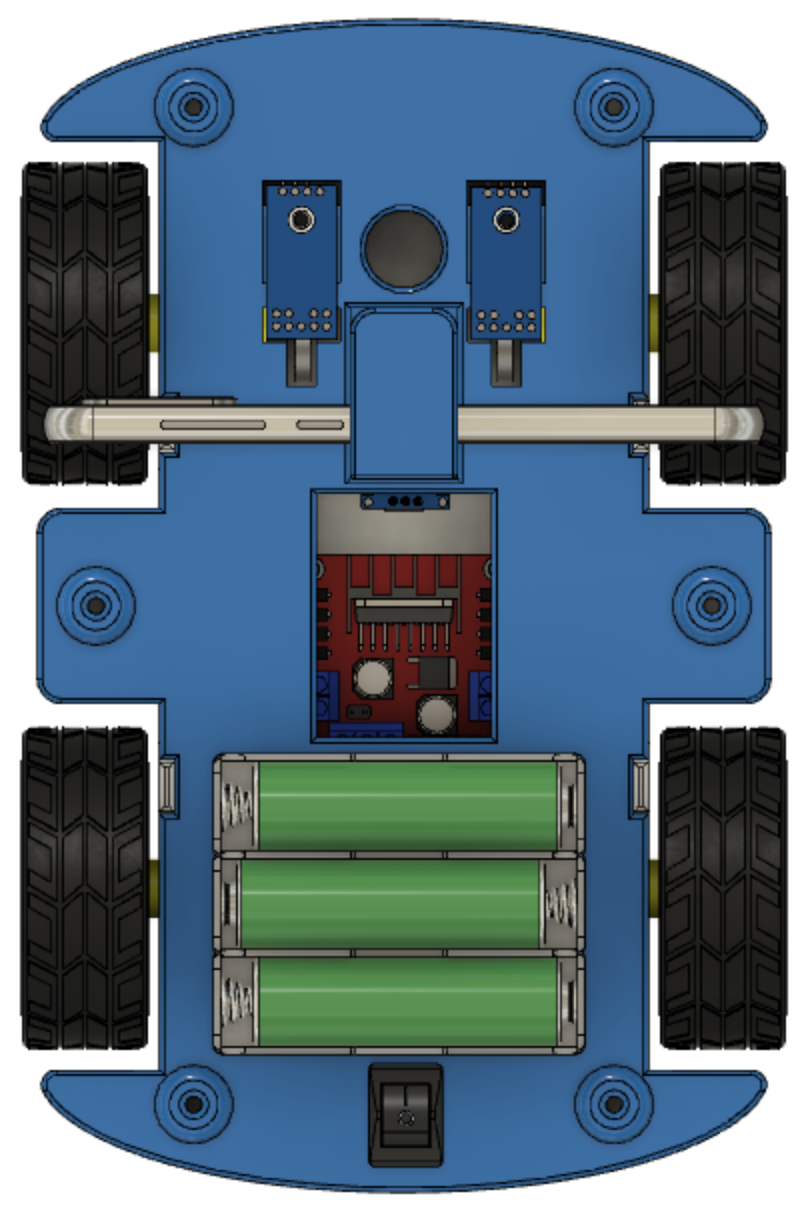}&
\includegraphics[height=3cm]{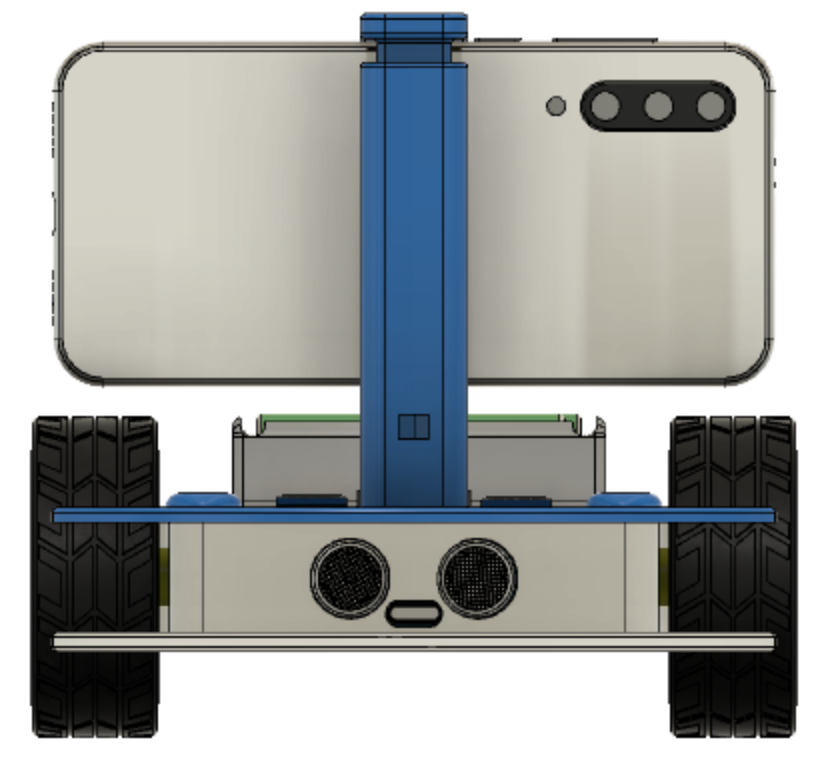}&
\includegraphics[height=3cm]{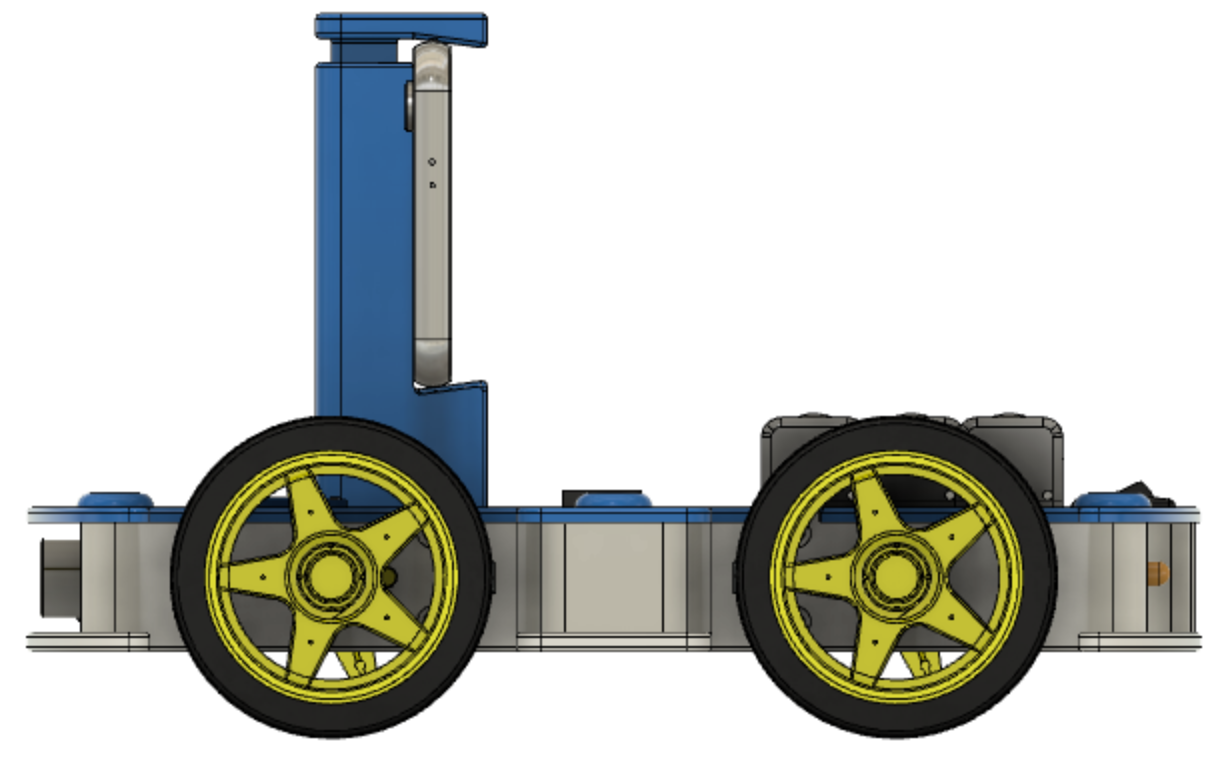}&
\includegraphics[height=3cm]{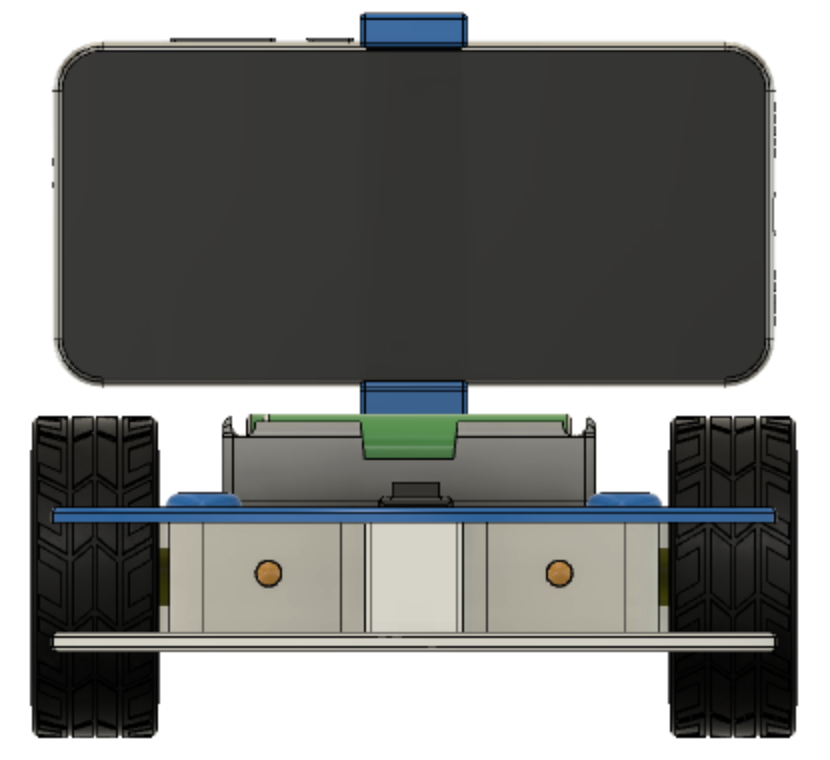}\\
\end{tabular}
\caption{\textbf{Mechanical design.} CAD design of the 3D-printed robot body.}
\label{fig:mechanial_design}
\end{figure*}

\subsection{A Body for a Low-cost Wheeled Robot}
A brain without a body cannot act. In order to leverage the computational power and sensing capabilities of a smartphone, the robot needs an actuated body. We develop a body for a low-cost wheeled robot which only relies on readily available electronics and 3D printing for its chassis. The total cost is \$50 for building a single body, with 30\% of that cost due to good batteries. If building multiple robots, the cost further decreases. 
\tblLabel~\ref{tbl:bom} shows the bill of materials.

\mypara{Mechanical design}
The chassis of the robot is 3D-printed and is illustrated in \figLabel \ref{fig:mechanial_design}. It consists of a bottom plate and a top cover which is fastened with six M3x25 screws. The bottom plate features the mounting points for the motors and electronics. The four motors are fixed with eight M3x25 screws. The motor controller and microcontroller attach to the bottom plate. There are appropriate openings for the indicator LEDs and the sonar sensor. The optical wheel odometry sensors can be mounted on the top cover and the bottom plate has grooves for the corresponding encoder disks on the front wheels. The top plate features a universal smartphone mount which uses a spring to adjust to different phones. There is also an opening for the USB cable that connects the smartphone to an Arduino microcontroller.

With standard settings on a consumer 3D printer (\eg Ultimaker S5), the complete print requires 13.5 hours for the bottom plate and 9.5 hours for the top cover with the phone mount. It is possible to print at a faster pace with less accuracy. The material weight is 146g for the bottom and 103g for the top. Considering an average PLA filament price of \$20/kg, the total material cost is about \$5.

\begin{figure}
 \centering
 \includegraphics[width=\linewidth]{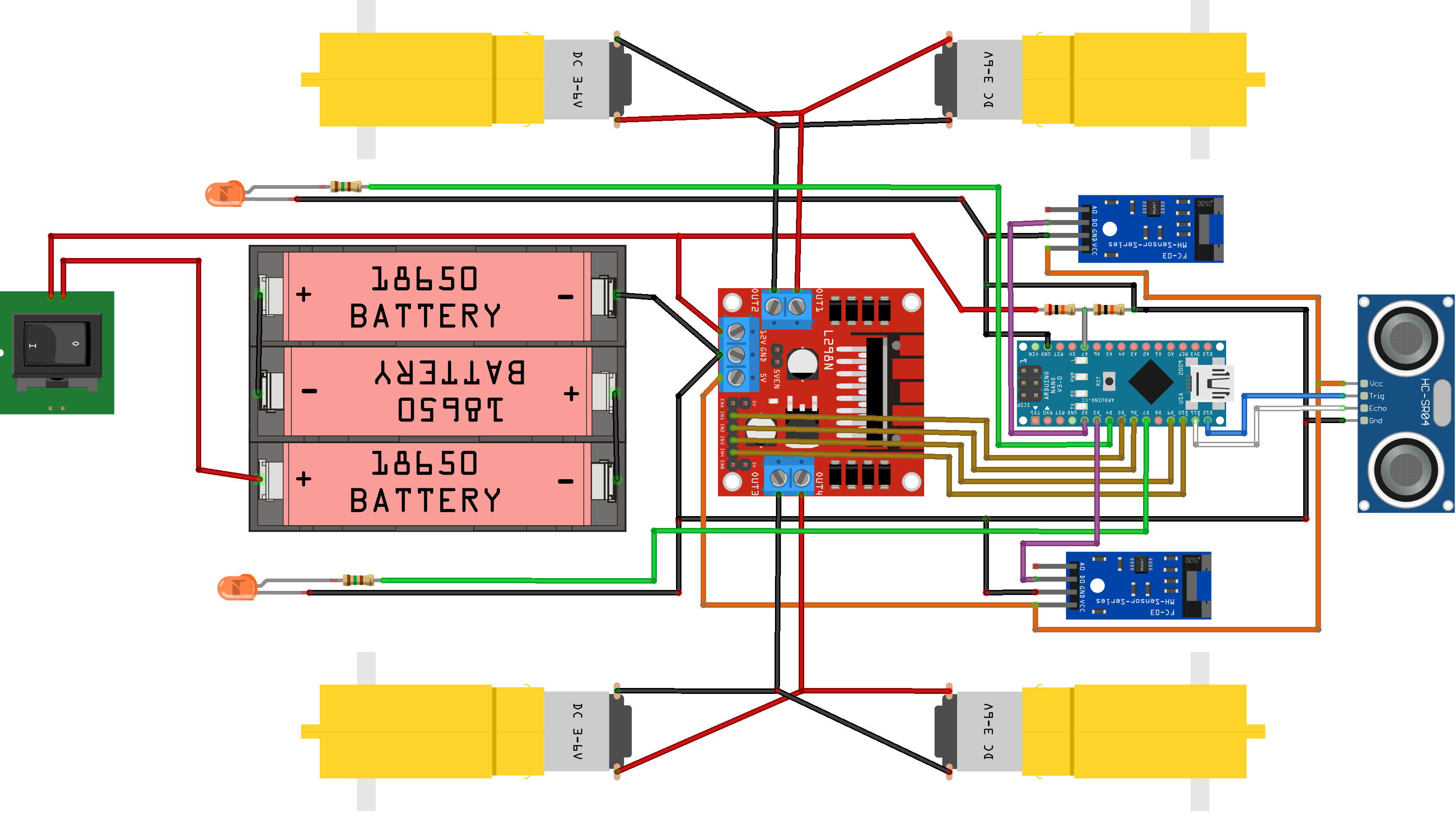}\\
 \vspace{8pt}
 \includegraphics[width=\linewidth]{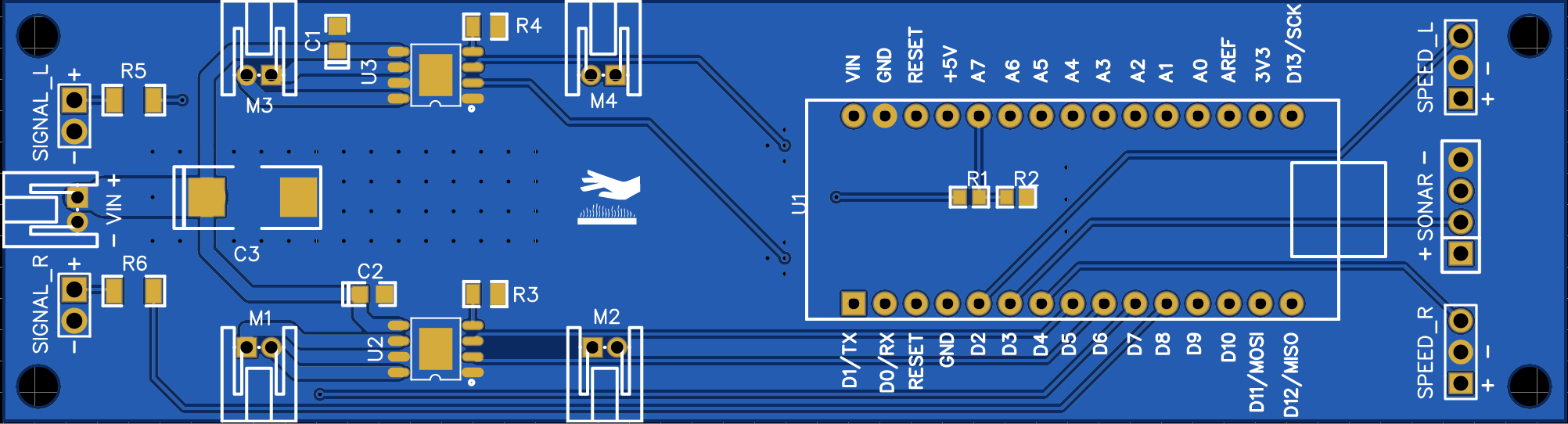}
 \vspace{-8pt}
 \caption{\textbf{Wiring diagram.} \textit{Top:} Electrical connections between battery, motor driver, microcontroller, speed sensors, sonar sensor and indicator LEDs. \textit{Bottom:} Optional custom PCB to reduce wiring.}
 \label{fig:electrical_design}
\vspace{-8pt}
\end{figure}

\mypara{Electrical design}
\figLabel \ref{fig:electrical_design} shows the electrical design. We use the L298N breakout board as motor controller. The two left motors are connected to one output and the two right motors to the other.
The battery pack is connected to the power terminals to provide power to the motors as needed.
Our battery consists of three USB-rechargeable 18650 Lithium cells connected in series, providing a voltage between 9.6V and 12.6V depending on their state-of-charge (SOC). An Arduino Nano board is connected to the smartphone via its USB port, providing a serial communication link and power. Two LM393-based speed sensors, a sonar sensor and two indicator LEDs are connected to six of the digital pins. The optical speed sensors provide a wheel odometry signal for the left and right front wheels.  
The sonar sensor provides a coarse estimate of free space in front of the robot. The indicator LEDs can be toggled, providing visual means for the robot to communicate with its environment. We also use one of the analog pins as input to measure the battery voltage through a voltage divider. Finally, four PWM pins are connected to the motor controller. This allows us to adjust the speed and direction of the motors according to the control commands received from the smartphone. We have also designed a PCB with integrated battery monitoring and two TI-DRV8871 motor drivers for increased efficiency. The Arduino, motors, indicator LEDs, speed sensors, and the ultrasonic sensor are simply plugged in. When building multiple robots, the PCB further reduces setup time and cost.

\subsection{Software Stack}
\begin{figure}[!htb]
 \vspace{4pt}
 \centering
 \includegraphics[trim=40 50 150 110, clip, width=\columnwidth]{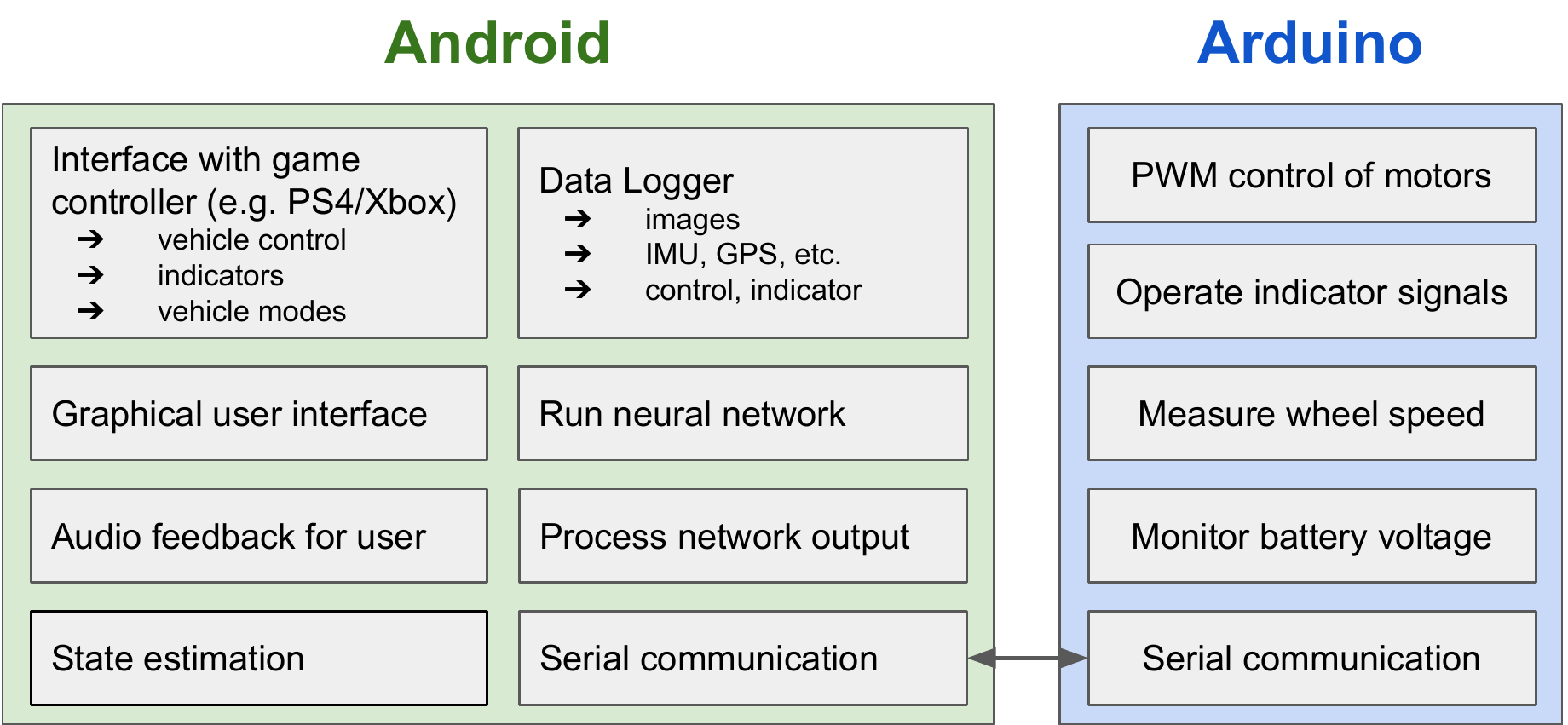}
 \vspace{-8pt}
 \caption{\textbf{Software design.} Our Android application is responsible for high-level computation on the smartphone and the Arduino program provides the low-level interface to the vehicle.}
 \label{fig:software_design}
 \vspace{-8pt}
\end{figure}

Our software stack consists of two components, illustrated in \figLabel \ref{fig:software_design}. The first is an Android application that runs on the smartphone. Its purpose is to provide an interface for the operator, collect datasets, and run the higher-level perception and control workloads. The second component is a program that runs on the Arduino. It takes care of the low-level actuation and some measurements such as wheel odometry and battery voltage. The Android application and the Arduino communicate via a serial communication link. In the following, we discuss both components in more detail.

\mypara{Android application}
We design a user interface which provides visual and auditory feedback for interaction with the robot. We use Bluetooth communication to interface with common game console controllers (\eg PS4, Xbox), which can be used to teleoperate the robot for data collection. 
To collect data, such as demonstrations for imitation learning, we use the joystick inputs to control the robot and use the buttons to trigger functionalities such as toggling control modes, logging, running a neural network, \etc We derive our graphical user interface from the Android Tensorflow Object Detection application \cite{tf_od_app} and extend it. Our GUI provides the camera feed and buttons to toggle data logging, control modes, and serial communication. It also allows switching between different neural networks to control the vehicle and provides relevant information such as image resolution, inference time and predicted controls. We also integrate voice feedback for operation via the game controller. 

The Android ecosystem provides a unified interface to obtain sensor data from any Android smartphone. We build a data logger on top of that in order to collect datasets with the robots. Currently, we record readings from the following sensors: camera, gyroscope, accelerometer, magnetometer, ambient light sensor, and barometer. Using the Android API, we are able to obtain the following sensor readings: RGB images, angular speed, linear acceleration, gravity, magnetic field strength, light intensity, atmospheric pressure, latitude, longitude, altitude, bearing, and speed.
In addition to the phone sensors, we also record body sensor readings (wheel odometry, free space estimate and battery voltage), which are transmitted via the serial link.

We leverage the computational power of the smartphone to process the sensory input and compute the robots' actions in real time. While there are many classic motion planning algorithms, we focus on learning-based approaches, which allow for a unified interface. In particular, we rely on the Tensorflow Lite infrastructure, which integrates seamlessly with smartphones~\cite{ignatov2019ai,lee2019device}. Our Android application features model definitions for object detection and autonomous navigation. These define the input and output properties of the neural network. 
We build on top of the Tensorflow Object Detection application \cite{tf_od_app} to detect people and perform visual servoing to follow them. We also integrate a model for autonomous navigation inspired by Conditional Imitation Learning \cite{codevilla2018end}. The deployment process is simple. After training a model in Tensorflow, it is converted to a Tensorflow Lite model that can be directly deployed on the smartphone. Both neural networks only rely on the camera feed for their predictions. Neither wheel odometry nor sonar sensor readings are required to produce the raw vehicle controls.

\mypara{Arduino program}
We use an Arduino Nano microcontroller to act as a bridge between the vehicle body and the smartphone. Its main task is to handle the low-level control of the vehicle and provide readings from vehicle-mounted sensors.
The program components are shown on the right in \figLabel~\ref{fig:software_design}. The Arduino receives the vehicle controls and indicator signals via the serial connection. It converts the controls to PWM signals for the motor controller and toggles the LEDs according to the indicator signal. The Arduino also keeps track of the wheel rotations by counting the interrupts at the optical sensors on the left and right front wheels. The sonar sensor transmits and receives a sound wave and the duration is measured and used to estimate the free space in front of the vehicle.
The battery voltage is calculated by a scaled moving average of measurements at the voltage divider circuit. All measurements are sent back to the Android application through the serial link.

\begin{table*}[!htb]
    \vspace{4pt}
    \setlength{\tabcolsep}{2.5pt} 
    \centering
	\begin{tabular}{l|ccccccccccccccccccccc}
	    \toprule
		\textbf{Platform}  & \rot{Retail Cost} & \rot{Setup Time [h]} & \rot{Size [cm]} & \rot{Weight [kg]} & \rot{Speed [m/s]} & \rot{Battery [min]} & \rot{Actuation} & \rot{Odometry} & \rot{Camera} & \rot{LiDAR} & \rot{Sonar} & \rot{IMU} & \rot{GPS} & \rot{WiFi} & \rot{Bluetooth} & \rot{3G/4G/5G}  & \rot{Speaker} & \rot{Microphone} & \rot{Display} & \rot{Compute} & \rot{Ecosystem}\\ 
		\midrule
		AutoRally~\cite{goldfain2019autorally}& \$10,000 & 100 & 100x60x40 & 22 & 25 & 20+ & BLDC+Servo & \YesV & \YesV & \NoX & \NoX & \YesV & \YesV & \YesV & \NoX & \NoX & \NoX & \NoX & \NoX &  Mini-ITX PC & ROS \\ 
		F1/10~\cite{o2019f1}                  & \$3600 & 3 & 55x30x20 & 4.5 & 18 & 20+ & BLDC+Servo & \YesV & \YesV & \YesV & \NoX & \YesV & \NoX & \YesV & \YesV & \NoX & \NoX & \NoX & \NoX & Jetson TX2 & ROS\\ 
		RACECAR~\cite{karaman2017project}     & \$3400 & 10 & 55x30x20 & 4.5 & 18 & 20+ & BLDC+Servo & \YesV & \YesV & \YesV & \NoX & \YesV & \NoX & \YesV & \YesV & \NoX & \NoX & \NoX & \NoX & Jetson TX1 & ROS\\
		BARC~\cite{gonzales2016autonomous}    & \$1000 & 3 & 54x28x21 & 3.2 & - & 20+ & BLDC+Servo & \YesV & \YesV & \NoX & \NoX & \YesV & \NoX & \YesV & \NoX & \NoX & \NoX & \NoX & \NoX & Odroid XU-4 & ROS \\ 
		MuSHR~\cite{srinivasa2019mushr}       & \$900 & 3 & 44x28x14 & 3 & 11 & 20+ & BLDC+Servo & \YesV & \YesV & \YesV & \NoX & \YesV & \NoX & \YesV & \YesV & \NoX & \NoX & \NoX & \NoX & Jetson Nano & ROS\\ 
		DeepRacer~\cite{balaji2019deepracer}  & \$400 & 0.25 & - & - & 6 & 15+ & BDC+Servo & \NoX & \YesV & \NoX & \NoX & \YesV & \NoX & \YesV & \NoX & \NoX & \NoX & \NoX & \NoX & Intel Atom & Custom\\
		DonkeyCar~\cite{donkey_car}           & \$250 & 2 & 25x22x12 & 1.5 & 9 & 15+ & BDC+Servo & \NoX & \YesV & \NoX & \NoX &\NoX & \NoX& \NoX&\NoX & \NoX & \NoX & \NoX & \NoX & Raspberry Pi & Custom \\ 
		\midrule
		Duckiebot~\cite{paull2017duckietown}  & \$280 & 0.5 & - & - & - & - & 2xBDC & \NoX & \YesV & \NoX & \NoX & \NoX & \NoX & \YesV & \NoX & \NoX & \NoX & \NoX & \NoX & Raspberry Pi & Custom\\ 
		Pheeno~\cite{wilson2016pheeno}        & \$270 & - & 13x11 & - & 0.42 & 300+ & 2xBDC & \YesV & \NoX & \NoX & \NoX & \YesV & \NoX & \YesV & \YesV & \NoX & \NoX & \NoX & \NoX & ARM Cortex-A7 & Custom\\ 
        JetBot~\cite{nvidia_jetbot}           & \$250 & 1 & 20x13x13 & - & - & - & 2xBDC & \NoX & \YesV & \NoX & \NoX & \NoX & \NoX & \YesV & \NoX & \NoX & \NoX & \NoX & \NoX & Nvidia Jetson & Custom\\ 
		Create-2~\cite{dekan2013irobot}       & \$200 & - & 34x34x9 & 3.6 & - & - & 2xBDC & \NoX & \NoX & \NoX & \NoX & \NoX & \NoX & \NoX & \NoX & \NoX & \NoX & \NoX & \NoX & \NoX & Custom\\ 
		Thymio II~\cite{riedo2013thymio}      & \$170 & - & 11x11x5 & 0.46 & 0.14 & - & 2xBDC & \NoX & \NoX & \NoX & \NoX & \YesV & \NoX & \NoX & \NoX & \NoX & \YesV & \YesV & \NoX & Microcontroller & Custom\\ 
		AERobot~\cite{rubenstein2015aerobot}  & \$20 & 0.1 & 3x3x3 & 0.03 & - & - & 2xVibration & \NoX & \NoX & \NoX & \NoX & \NoX & \NoX & \NoX & \NoX & \NoX & \NoX & \NoX & \NoX & Microcontroller & Custom\\ 
        \midrule
		\textbf{OpenBot}					      & \$50$^\star$ & 0.25 & 24x15x12 & 0.7 & 1.5 & 45+ & 4xBDC & \YesV & \YesV & \NoX & \YesV & \YesV & \YesV & \YesV & \YesV & \YesV & \YesV & \YesV & \YesV & Smartphone & Android \\ 
		\bottomrule
	\end{tabular}
	\caption{\textbf{Robots.} Comparison of wheeled robotic platforms. Top: Robots based on RC trucks. Bottom: Navigation robots for deployment at scale and in education. "--" indicates that no information is available. *The cost of the smartphone is not included and varies. }
	\label{tbl:platform_comparison}
	\vspace{-12pt}
\end{table*}

\subsection{Comparison to Other Wheeled Robots}

We compare to existing robot platforms in \tblLabel~\ref{tbl:platform_comparison}. In contrast to other robots, our platform has an abundance of processing power, communication interfaces, and sensors provided by the smartphone. Existing robots often rely on custom software ecosystems, which require dedicated lab personnel who maintain the code, implement new features, and implement drivers for new sensors. In contrast, we use Android, one of the largest constantly evolving software ecosystems. All the low-level software for sensor integration and processing already exists and improves without any additional effort by the robotics community. All sensors are already synchronized on the same clock, obviating what is now a major challenge for many existing robots.

\section{Evaluation}
\label{sec:evaluation}

To demonstrate that smartphones are suitable to provide sensing, communication, and compute for interesting robotics applications, we evaluate the presented platform on two applications, person following and autonomous navigation. The experimental setups are described in \secLabel \ref{sec:person_following} and in \secLabel \ref{sec:autonomous_navigation} respectively. We conduct the evaluation using a variety of popular smartphones from the past two years with prices ranging from \$120 to \$750. The smartphones are carefully selected to cover different manufactures, chipsets, and sensor suites. Detailed specifications and benchmark scores of the smartphones are provided in \secLabel \ref{sec:smartphones}.
We discuss our general evaluation setup and procedure that ensures a fair comparison in the following and report the results in Section~\ref{sec:results}.

\subsection{Evaluation metrics}
In order to streamline our evaluation while providing a comprehensive performance summary, we use three metrics: distance, success rate, and collisions. The distance is continuous and we report it as a percentage of the complete trajectory. The distance measurement stops if an intersection is missed, a collision occurs or the goal is reached. The success rate is binary and indicates whether or not the goal was reached. We also count the number of collisions. All results are averaged across three trials.  

\subsection{Evaluation protocol}
Since our experiments involve different smartphones, cheap robots, and a dynamic physical world, we make several considerations to ensure a fair evaluation.
We divide each experiment into several well-defined segments to ensure consistency and minimize human error. To ensure that the robots are initialized at the same position for each experiment, we use markers at the start and end position of each segment. We also align all phones with their power button to the phone mount to ensure the same mounting position across experiments. Since the inference time of smartphones can be affected by CPU throttling, we check the temperature of each smartphone before starting an experiment and close all applications running in the background. We use several metrics to provide a comprehensive performance analysis.

\section{Results}
\label{sec:results}

\subsection{Person Following}
In this experiment, we investigate the feasibility of running complex AI models on smartphones. We use object detectors and apply visual servoing to follow a person. Our experiments show that all recent mid-range smartphones are able to track a person consistently at speeds of 10 fps or higher. The cheapest low-end phone (Nokia 2.2) performs worst, but is surprisingly still able to follow the person about half of the time. We expect that even low-end phones will be able to run complex AI models reliably in the near future. The Huawei P30 Pro was the best performer in our comparison, closely followed by other high-end phones such as the Google Pixel 4XL and the Xiaomi Mi9. All recent mid-range phones (\eg Xiaomi Note 8, Huawei P30 Lite, Xiaomi Poco F1) clearly outperform the Samsung Galaxy Note 8, which was a high-end phone just two years ago. This is due to dedicated AI accelerators present in recent smartphones \cite{ignatov2019ai} and highlights the rapid rate at which smartphones are improving. Please see the supplementary video for qualitative results.

\begin{table}
\vspace{4pt}
\setlength{\tabcolsep}{3pt}
\centering
\begin{tabular}{@{}l|ccccrccc@{}}
\toprule
& \multicolumn{2}{c}{Distance $\uparrow$} & \multicolumn{2}{c}{Success $\uparrow$} & \multicolumn{2}{c}{Collisions $\downarrow$} & \multicolumn{2}{c}{FPS $\uparrow$} \\ 
MobileNet &V1 & V3 & V1  & V3 & ~~~V1& ~V3~~& V1 & V3\\
\midrule
Huawei P30 Pro      & 100\% & 100\% & 100\% & 100\% & 0.0 & 0.0 & 33 & 30\\
Google Pixel 4XL    & 100\% & 100\% & 100\% & 100\% & 0.0 & 0.0 & 32 & 28\\
Xiaomi Mi9          & 100\% & 100\% & 100\% & 100\% & 0.0 & 0.0 & 32 & 28 \\
Samsung Note 10     & 100\% & 100\% & 100\% & 100\% & 0.0 & 0.0 & 16 & 22\\
OnePlus 6           & 100\% & 100\% & 100\% & 100\% & 0.0 & 0.0 & 11 & 15\\
Huawei P30 Lite     & 100\% & 99\%  & 100\% & 83\%  & 0.0 & 0.3 & 9  & 11\\
Xiaomi Note 8       & 100\% & 100\% & 100\% & 100\% & 0.0 & 0.0 & 9  & 11 \\
Xiaomi Poco F1         & 98\%  & 100\% & 83\%  & 100\% & 0.3 & 0.0 & 8  & 12 \\
Samsung Note 8      & 58\%  & 100\% & 33\%  & 100\% & 0.0 & 0.0 & 6  & 10 \\
Nokia 2.2           & 37\%  & 50\%  & 0\%   & 0\%   & 0.0 & 0.3 & 4  & 5 \\
\bottomrule
\end{tabular} 
\caption{\textbf{Person following.} We use MobileNet detectors and visual servoing to follow a person. All results are averaged across three trials.}
\vspace{-8pt}
\end{table}

\subsection{Autonomous Navigation}
We train a driving policy that runs in real time on most smartphones. Our learned policy is able to consistently follow along corridors and take turns at intersections. We compare it to existing driving policies and achieve similar performance as the baselines while requiring about one order of magnitude fewer parameters. We also successfully transfer our driving policy to different smartphones and robot bodies. When training on data acquired with multiple smartphones and robots, we observe increased robustness. We show that our driving policy is able to generalize to previously unseen environments, novel objects, and even dynamic obstacles such as people even though only static obstacles were present in the training data.

\mypara{Comparing driving policies}
OpenBot enables benchmarking using real robots. We compare our policy to two baselines across three trials in \tblLabel~\ref{tbl:baseline_exp}.  
To ensure optimal conditions for the baselines, we use the high-end smartphone Xiaomi Mi9. Our driving policy network is smaller by a factor of 7 or more than the baselines. Yet it outperforms PilotNet~\cite{bojarski2016end} and achieves similar performance to CIL~\cite{codevilla2018end} while running at twice the speed. 

\begin{table}[!htb]
\setlength{\tabcolsep}{3pt}
\centering
\begin{tabular}{l|ccccc}
\toprule
& Distance $\uparrow$ & Success $\uparrow$ & Collisions $\downarrow$ & FPS $\uparrow$ & Params $\downarrow$\\
\midrule
PilotNet \cite{bojarski2016end}   & 92$\pm$0\% & 83$\pm$0\% & 0.0$\pm$0.0 & 60$\pm$1 & 9.6M \\
CIL \cite{codevilla2018end}     & 94$\pm$5\% & 89$\pm$10\% & 0.0$\pm$0.0 & 20$\pm$1 & 10.7M\\
Ours                            & 94$\pm$5\% & 89$\pm$10\% & 0.0$\pm$0.0 & 47$\pm$2 & 1.3M\\
\bottomrule
\end{tabular} 
\caption{\textbf{Baselines.} We compare our driving policy to two baselines. All policies are trained for 100 epochs using the same data and hyperparameters.}
\label{tbl:baseline_exp}
\vspace{-12pt}
\end{table}

\mypara{Generalization to novel phones}
\tblLabel \ref{tbl:transfer_phone} shows that our navigation policy can be trained with data from one phone and generalize to other phones. How well the generalization works depends on the target phone, especially its processing power and camera placement. 
We observe a degradation in performance for phones unable to run the driving policy in real time.
Differences in camera placement affect qualitative driving performance; for tasks that require high precision this may need to be accounted for. The differences in camera sensors (\eg color reproduction and exposure) are largely overcome by data augmentation.

\begin{table}[!htb]
\vspace{4pt}
\setlength{\tabcolsep}{4pt}
\centering
\begin{tabular}{l|clccc}
\toprule
& Distance $\uparrow$ & Success $\uparrow$ & Collisions $\downarrow$ & FPS $\uparrow$\\
\midrule
Xiaomi Mi9   & 94$\pm$5\% & 89$\pm$10\% & 0.0$\pm$0.0 & 47$\pm$2\\
\midrule
Google Pixel 4XL   & 92$\pm$0\% & 83$\pm$0\% &  0.0$\pm$0.0 & 57$\pm$3 \\
Huawei P30 Pro   & 97$\pm$5\% & 94$\pm$10\% & 0.0$\pm$0.0 & 51$\pm$0\\
Samsung Note 10   & 92$\pm$0\% & 83$\pm$0\% &  0.0$\pm$0.0 & 38$\pm$8\\
OnePlus 6T   & 89$\pm$5\% & 78$\pm$10\% & 0.1$\pm$0.1 & 32$\pm$0\\
Xiaomi Note 8   & 92$\pm$0\% & 83$\pm$0\% &  0.0$\pm$0.0 & 31$\pm$0\\
Huawei P30 Lite   & 92$\pm$0\% & 83$\pm$0\% &  0.0$\pm$0.0 & 30$\pm$1\\
Xiaomi Poco F1   & 86$\pm$5\% & 72$\pm$10\% &  0.1$\pm$0.1 & 26$\pm$8\\
Samsung Note 8   & 83$\pm$0\% & 67$\pm$0\% &  0.2$\pm$0.0 & 19$\pm$3\\
\bottomrule
\end{tabular} 
\caption{\textbf{Novel phones.} We train our driving policy using one phone (top) and then test it on other phones (bottom).}
\label{tbl:transfer_phone}
\vspace{-12pt}
\end{table}

\mypara{Generalization to novel bodies}
\tblLabel \ref{tbl:transfer_body} shows that our navigation policy can be trained with data from one robot body and generalize to other robot bodies. Due to the cheap components, every body exhibits different actuation noise which may change over time and is observable in its behaviour (\eg a tendency to pull to the left or to the right). We address this by injecting noise in the training process \cite{codevilla2018end}. Further details are provided in \secLabel \ref{sec:dataset}.

\begin{table}[!htb]
\setlength{\tabcolsep}{9pt}
\centering
\begin{tabular}{l|ccccc}
\toprule
& Distance $\uparrow$ & Success $\uparrow$ & Collisions $\downarrow$ \\
\midrule
Robot Body 1   & 94$\pm$5\% & 89$\pm$10\% & 0.0$\pm$0.0 \\
\midrule
Robot Body 2   & 94$\pm$5\% & 89$\pm$10\% & 0.0$\pm$0.0 \\
Robot Body 3   & 92$\pm$0\% & 83$\pm$0\% & 0.0$\pm$0.0 \\
Robot Body 4   & 89$\pm$5\% & 78$\pm$10\% & 0.1$\pm$0.1 \\
\bottomrule
\end{tabular} 
\caption{\textbf{Novel bodies.} We train our driving policy using one body (top) and then test it on other bodies (bottom).}
\label{tbl:transfer_body}
\vspace{-16pt}
\end{table}

\mypara{Generalization to novel obstacles}
Even though our driving policies were only exposed to static obstacles in the form of office chairs during data collection, they were able to generalize to novel static obstacles (potted plants) and even dynamic obstacles (people) at test time.
The low image resolution, aggressive downsampling, and small number of parameters in our network may serve as natural regularization that prevents the network from overfitting to specific obstacles. Since the network processes camera input on a frame-by-frame basis, static and dynamic obstacles are treated on the same basis. We also conjecture that the network has learned some robustness to motion blur due to vibrations of the vehicle. Our navigation policy is also able to generalize to novel environments within the same office building.
Please refer to the supplementary video for qualitative results.

\mypara{Learning with multiple robots}
We also investigated the impact of using multiple different smartphones and robot bodies for data collection which is relevant for using our platform at scale. We provide detailed results in \secLabel \ref{sec:additional_experiments} 
and summarize the findings here.
Training the driving policies on data acquired from multiple smartphones improves generalization to other phones; every manufacturer tunes the color reproduction and exposure slightly differently, leading to natural data augmentation.
The driving policy trained on data acquired with multiple robot bodies is the most robust; since the smartphone was fixed, the network had to learn to cope with noisy actuation and dynamics, which we show to be possible even with relatively small datasets.

\section{Conclusion} 
\label{sec:conclusion}

This work aims to address two key challenges in robotics: accessibility and scalability.
Smartphones are ubiquitous and are becoming more powerful by the year. We have developed a combination of hardware and software that turns smartphones into robots. The resulting robots are inexpensive but capable.
Our experiments have shown that a \$50 robot body powered by a smartphone is capable of person following and real-time autonomous navigation.
We hope that the presented work will open new opportunities for education and large-scale learning via thousands of low-cost robots deployed around the world.

Smartphones point to many possibilities for robotics that we have not yet exploited. For example, smartphones also provide a microphone, speaker, and screen, which are not commonly found on existing navigation robots. These may enable research and applications at the confluence of human-robot interaction and natural language processing. We also expect the basic ideas presented in this work to extend to other forms of robot embodiment, such as manipulators, aerial vehicles, and watercrafts.

\section*{APPENDIX}

\section{System Overview}
\label{sec:system_overview}

\begin{figure}[!htb]
\centering
 \includegraphics[trim=40 30 100 90, clip, width=\columnwidth]{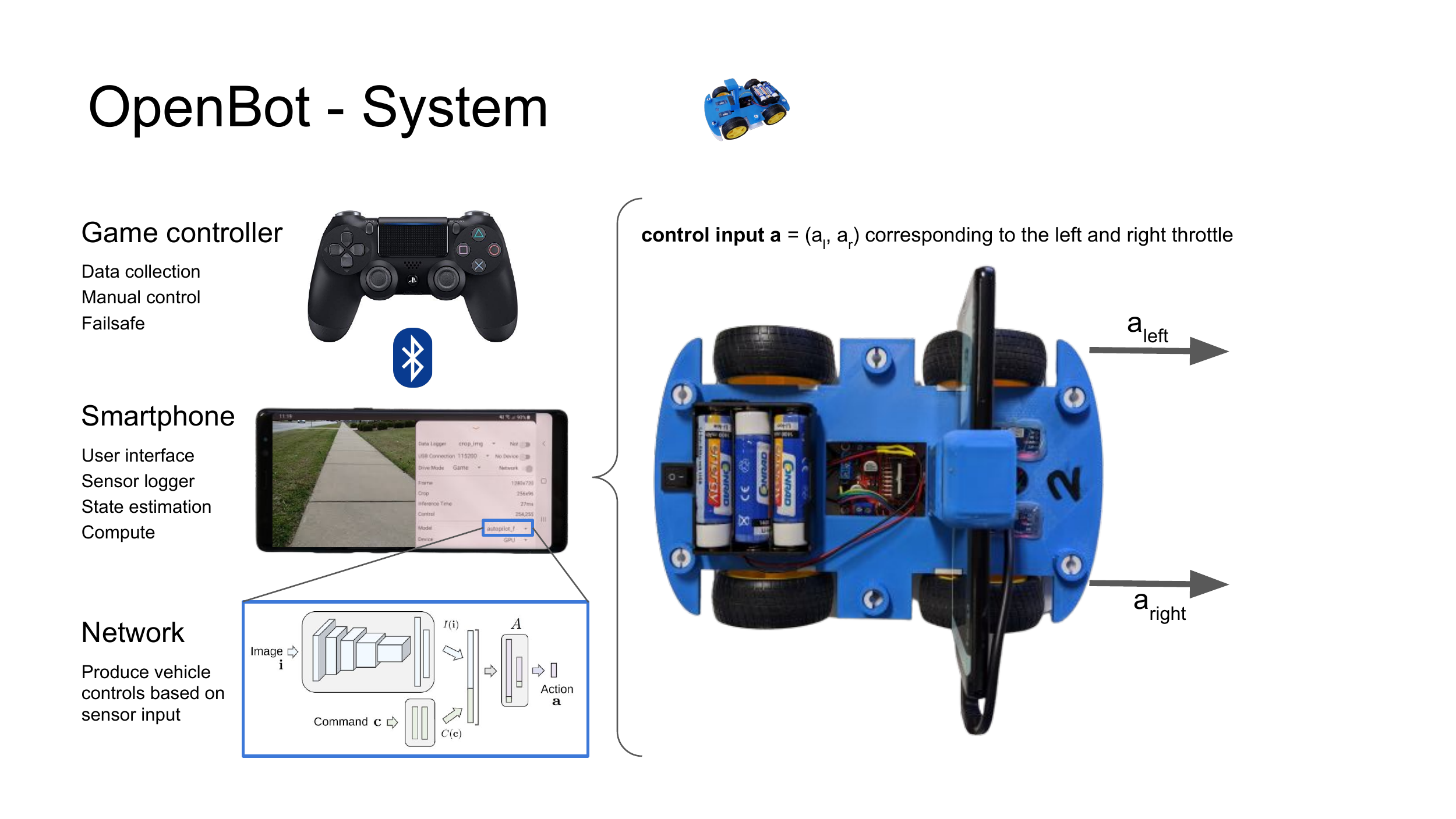}
 \caption{\textbf{System overview.} Our wheeled robot leverages a smartphone for sensing and computation. The robot body costs \$50 without the smartphone. The platform supports person following and real-time autonomous navigation in unstructured environments.}
\label{fig:system}
\end{figure}

\figLabel \ref{fig:system} depicts the high-level overview of our system. It comprises a smartphone mounted onto a low-cost robot body. The smartphone consumes sensor data (\eg images, IMU, GPS, \etc) and optionally user input in order to produce high-level controls for the vehicle such as steering angle and throttle. The microcontroller on the robot body applies the corresponding low-level actuation signals to the vehicle.

\begin{table*}[!htb]
\setlength{\tabcolsep}{3pt} 
\centering
\begin{tabular}{lcccccccccc}
\toprule
\textbf{Mobile Phone} & \textbf{Release} & \textbf{Price} & \textbf{Main Camera} & \textbf{Memory/RAM} & \textbf{CPU}& \textbf{GPU}& \textbf{Overall~} & \textbf{Graphics} & \textbf{AI} \\
\midrule
Samsung Note 8 & 09/17 & 300& 12 MP, f/1.7, 1/2.55" & 64GB, 6GB& Exynos 8895& Mali-G71 MP20& 3374 & 40890 & 4555 \\
Huawei P30 Pro & 03/19 & 650& 40 MP, f/1.6, 1/1.7" & 128GB, 8GB & HiSilicon Kirin 980 & Mali-G76 MP10 & 4654 & 45889 & 27112 \\
Google Pixel 4XL & 10/19 & 750& 12.2 MP, f/1.7, 1/2.55" & 64GB, 6GB & Qualcomm SM8150 & Adreno 640 & 5404 & -- & 32793 \\
Xiaomi Note 8 & 08/19 & 170 & 48 MP, f/1.8, 1/2.0" & 64GB, 4GB & Qualcomm SDM665 & Adreno 610 & 2923 & 17636 & 7908 \\
Xiaomi Mi 9 & 02/19 & 380 & 48 MP, f/1.8, 1/2.0" & 128GB, 6GB & Qualcomm SM8150 & Adreno 640 & 5074 & 45089 & 31725 \\
OnePlus 6T & 11/18 & 500 & 16 MP, f/1.7, 1/2.6" & 128GB, 8GB & Qualcomm SDM845 & Adreno 630 & 4941 & 43886 & 18500\\
\midrule
Samsung Note 10 & 08/19 & 750& 12 MP, f/1.5-2.4, 1/2.55" & 256GB, 8GB& Exynos 9825 & Mali-G76 MP12& 4544 & 45007 & 24924 \\
Huawei P30 Lite & 04/19 & 220& 48 MP, f/1.8, 1/2.0", & 128GB, 4GB & Hisilicon Kirin 710 & Mali-G51 MP4 & 2431 & 20560 & - \\
Xiaomi Poco F1 & 08/18 & 290& 12 MP, f/1.9, 1/2.55"& 64GB, 6GB & Qualcomm SDM845 & Adreno 630 & 4034 & 43652 & 6988\\
\midrule
Nokia 2.2& 06/19 & 120& 13 MP, f/2.2, 1/3.1" & 16GB, 2GB & Mediatek MT6761 & PowerVR GE8320 & 848 & 5669 & --\\ 
\bottomrule
\end{tabular}
\caption{\textbf{Smartphones.} Specifications of the smartphones used in our experiments. We report the overall, graphics, and AI performance according to standard benchmarks. Top: six smartphones used to collect training data. Bottom: smartphones used to test cross-phone generalization. "--" indicates that the score is not available. }
\label{tbl:smartphones}
\end{table*}

\section{Smartphones}
\label{sec:smartphones}

\tblLabel~\ref{tbl:smartphones} provides an overview of the smartphones we use in our experiments. We provide the main specifications along with the \textit{Basemark OS II} and \textit{Basemark X} benchmark scores which measure the overall and graphical performance of smartphones. We also include the AI score \cite{lee2019device} where available.

\section{Person Following}
\label{sec:person_following}

\subsection{Experimental setup}
The robot is tasked to follow a person autonomously. To this end, we run a neural network for object detection and only consider detections of the person class and reject detections with a confidence below a threshold of 50\%. We track detections across frames, and pick the one with the highest confidence as the target. We apply visual servoing with respect to the center of the bounding box, keeping the person centered in the frame. We evaluate two variants of the object detector (see \secLabel \ref{sec:object_detection_network}) across ten different smartphones. For fair comparison, we only use the CPU with one thread on each phone. Using the GPU or the NNAPI can further improve the runtime on most phones. We provide a quantitative evaluation in a controlled indoor environment. The route involves a round trip between an office and a coffee machine and includes four left turns and four right turns. We average results across three trials for each experiment. Please refer to \secLabel \ref{sec:results} for the results. In addition, the supplementary video contains qualitative results in unstructured outdoor environments.

\subsection{Network}
\label{sec:object_detection_network}
We use the SSD object detector with a pretrained MobileNet backbone~\cite{howard2017mobilenets}. To investigate the impact of inference time, we use two different versions of MobileNet, the original MobileNetV1~\cite{howard2017mobilenets} and the lastest MobileNetV3~\cite{howard2019searching}. We use the pretrained models released as part of the Tensorflow object detection API. Both models were trained on the COCO dataset~\cite{lin2014microsoft} with 90 class labels. The models are quantized for improved inference speed on smartphone CPUs. 

\section{Autonomous Navigation}
\label{sec:autonomous_navigation}

\subsection{Experimental setup}
The robots have to autonomously navigate through corridors in an office building without colliding. The driving policy receives high-level guidance in the form of indicator commands such as \textit{turn left / right at the next intersection}~\cite{codevilla2018end}.
Each trial consists of several segments with a total of 2 straights, 2 left turns, and 2 right turns. More details on the evaluation setup including a map with dimensions are provided in \secLabel \ref{sec:eval_details}.

\begin{figure}[!b]
 \centering
 \includegraphics[trim=10 50 20 20, clip, width=0.8\linewidth]{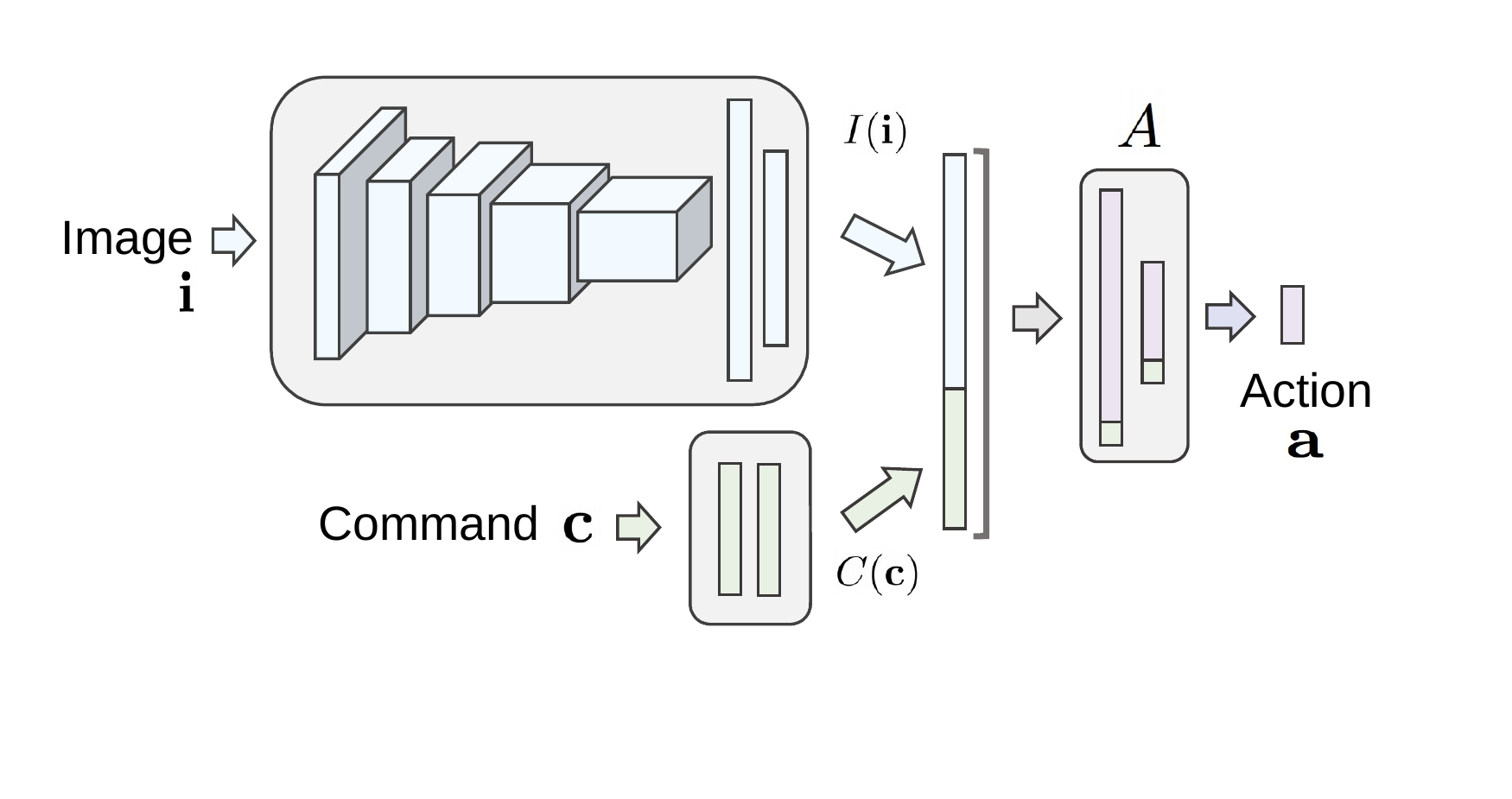}
 \caption{\textbf{Driving policy: Network architecture.} Our compact neural network for autonomous navigation runs in real time on most smartphones we tested.}
 \label{fig:network}
\end{figure}

\subsection{Network}
We design a neural network similar in spirit to the \textit{command-input} variant of Conditional Imitation Learning \cite{codevilla2018end}. Our network is about one order of magnitude smaller than existing networks and is able to run in real time even on mid-range smartphones. We train this network using a novel loss function (see \secLabel \ref{sec:training_process}) and validation metrics (see \secLabel \ref{sec:val_metrics}). We obtain successful navigation policies with less than 30 minutes of labeled data and augmentation. Our network is visualized in \figLabel \ref{fig:network}. It takes an image $\textbf{i}$ and a command $\textbf{c}$ as inputs and processes them via an image module $I(\textbf{i})$ and a command module $C(\textbf{c})$. The image module consists of five convolutional layers with $32$, $64$, $96$, $128$ and $256$ filters, each with a stride of $2$, a kernel size of $5$ for the first layer, and $3$ for the remaining layers. We apply \textit{relu} activation functions, \textit{batch-normalization}, and $20\%$ \textit{dropout} after each convolutional layer. The output is flattened and processed by two fully-connected layers with $128$ and $64$ units. The command module is implemented as an MLP with 16 hidden units and $16$ output units. The outputs of the image module and the command module are concatenated and fed into the control module $A$ which is also implemented as an MLP. It has two hidden layers with $64$ and $16$ units and then linearly regresses to the action vector $a$. We concatenate the command $\textbf{c}$ with the hidden units for added robustness. We apply $50\%$ \textit{dropout} after all fully-connected layers. 

We use an image input size of $256$x$96$, resulting in $1.3$M parameters. At the same input resolution, PilotNet \cite{bojarski2016end} has $9.6$M parameters and CIL \cite{codevilla2018end} has $10.7$M parameters. Our network runs in real time on most smartphones we tested. The average inference times on the Samsung Galaxy Note 10, Xiaomi Mi9, Xiaomi Pocofone F1, and Huawei P30 Lite are $19$ms, $21$ms, $29$ms, and $32$ms, respectively. Further speedups are possible by quantization of the network weights and by leveraging the GPU or the recent neural network API (NNAPI)~\cite{ignatov2019ai}.

\subsection{Dataset Collection}
\label{sec:dataset}
\begin{figure}[!htb]
\centering
 \includegraphics[trim=23 90 21 105, clip, width=\columnwidth]{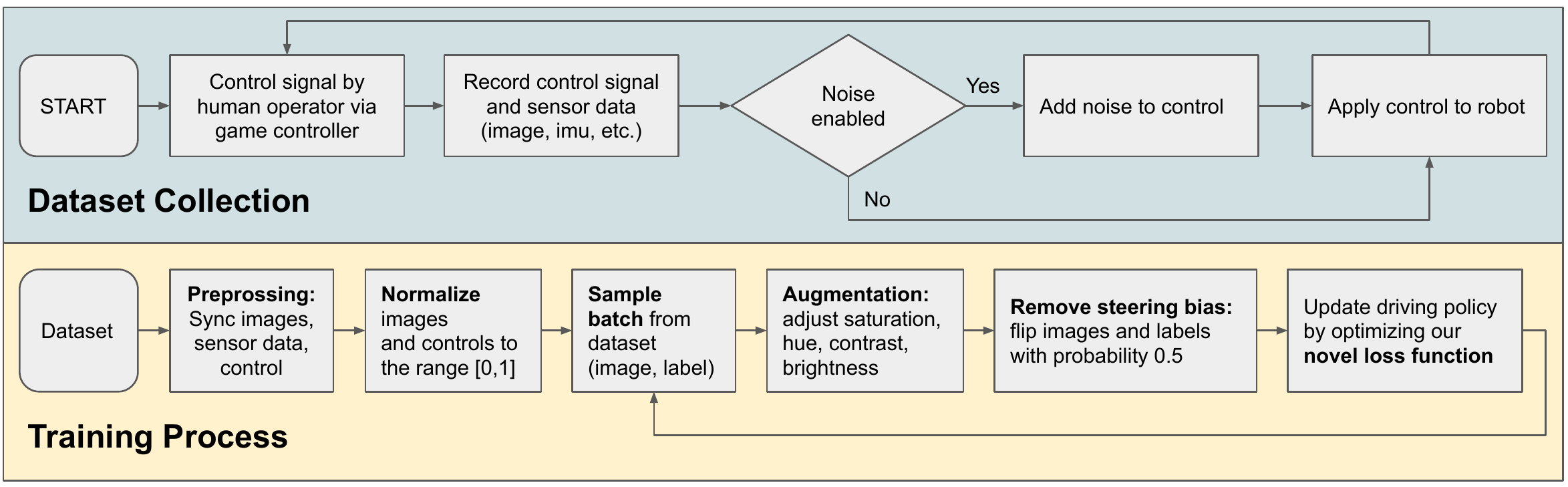}
 \caption{\textbf{Driving policy: Training pipeline.}  The flowchart explains the complete process for obtaining our autonomous navigation policy. There are two main components, dataset collection and training the driving policy which is represented by a neural network.}
\label{fig:training_pipeline}
\end{figure}

\figLabel \ref{fig:training_pipeline} depicts the pipeline for training our driving policy. We record a driving dataset with a human controlling the robot via a game controller. In previous works, data was often collected with multiple cameras for added exploration \cite{bojarski2016end,codevilla2018end}. Since we only use one smartphone camera, we inject noise during data collection and record the recovery maneuvers executed by the human operator \cite{codevilla2018end}. We also scatter obstacles in the environment, such as chairs, for added robustness. Please refer to \secLabel \ref{sec:training_env} for further details, including maps and images of the environment.

\subsection{Training Process}
\label{sec:training_process}
We augment the images by randomly adjusting hue, saturation, brightness and contrast during training. In addition, we flip images and labels to increase our effective training set size and balance potential steering biases. We normalize images and actions to the range $[0,1]$.

When training end-to-end driving policies on autonomous navigation datasets, one common challenge is the huge label imbalance. The majority of the time, the vehicle is driving in a straight line, resulting in many images with the same label. One common approach is to resample the dataset or carefully craft individual batches during training \cite{codevilla2018end}. However, this usually requires a fixed dataset or computational overhead. If the dataset is dynamically changing or arrives as a continuous stream these methods are not feasible. Instead, we address this imbalance with a weighted loss. The intuition is simple: the stronger the steering angle, the more critical the maneuver. Hence we use a loss with a weighted term proportional to the steering angle combined with a standard MSE loss on the entire action vector to ensure that throttle is learned as well: 

\begin{equation}
\mathbb{L} = w^2 \cdot \text{MSE} \left(s^t, s^p \right) +  \text{MSE} \left(\mathbf{a^t}, \mathbf{a^p}\right),
\end{equation}
where $\mathbf{a^t}$ is the target action, $\mathbf{a^p}$ is the predicted action, $s^t$ is the target steering angle, and $w = (s^t + b)$  with a bias $b$ to control the weight of samples with zero steering angle. Since our vehicle uses differential steering, the action vector consists of a two-dimensional control signal $\mathbf{a} = (a_l, a_r)$, corresponding to throttle for the left and right wheels. We compute the steering angle as $s = a_l-a_r$. 

We use the Adam optimizer with an initial learning rate of $0.0003$ and train all models for $100$ epochs. We obtain successful navigation policies with less than 30 minutes of labeled data and augmentation. Our validation metrics are further discussed in \secLabel \ref{sec:val_metrics}.

\subsection{Validation Metrics}
\label{sec:val_metrics}
A major challenge in training autonomous driving policies and evaluating them based on the training or validation loss is the lack of correlation to the final performance of the driving policy \cite{codevilla2018offline}. Different action sequences can lead to the same state. The validation loss measures the similarity between target and prediction which is too strict. Hence, we define two validation metrics which are less strict and reduce the gap between offline and online evaluation. The first metric measures whether the steering angle is within a given threshold, which we set to $0.1$. The second metric is even more relaxed and only considers whether the steering direction of the target and the prediction align. We find empirically that these metrics are more reliable as the validation loss. However, the correlation to the final driving performance is still weak. We pick the best checkpoint based on the average of these two metrics on a validation set. 

\subsection{Training environment}
\label{sec:training_env}

\begin{figure}[!htb]
 \centering
 \includegraphics[trim=10 130 10 100, clip, width=\linewidth]{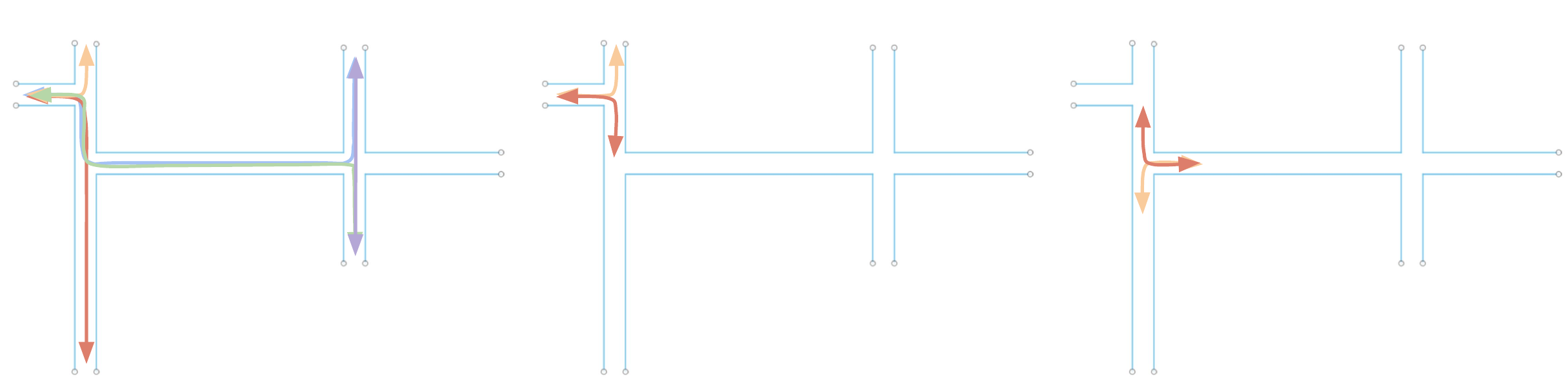}
 \caption{\textbf{Training Routes.} We collect data on three different routes: \textit{R1}, \textit{R2} and \textit{R3} (from left to right). \textit{R1} is composed of 5 bi-directional segments with a total of $20$ intersections: $8$ left turns, $8$ right turns, and $4$ straights. \textit{R2} and \textit{R3} are two different T-junctions each with two bi-directional segments with a total of two right turns, and two left turns.}
 \label{fig:training_routes}
\end{figure}

\begin{figure}[!htb]
\centering
\setlength{\tabcolsep}{2pt}
 \begin{tabular}{cccc}
      \includegraphics[width=0.24\columnwidth]{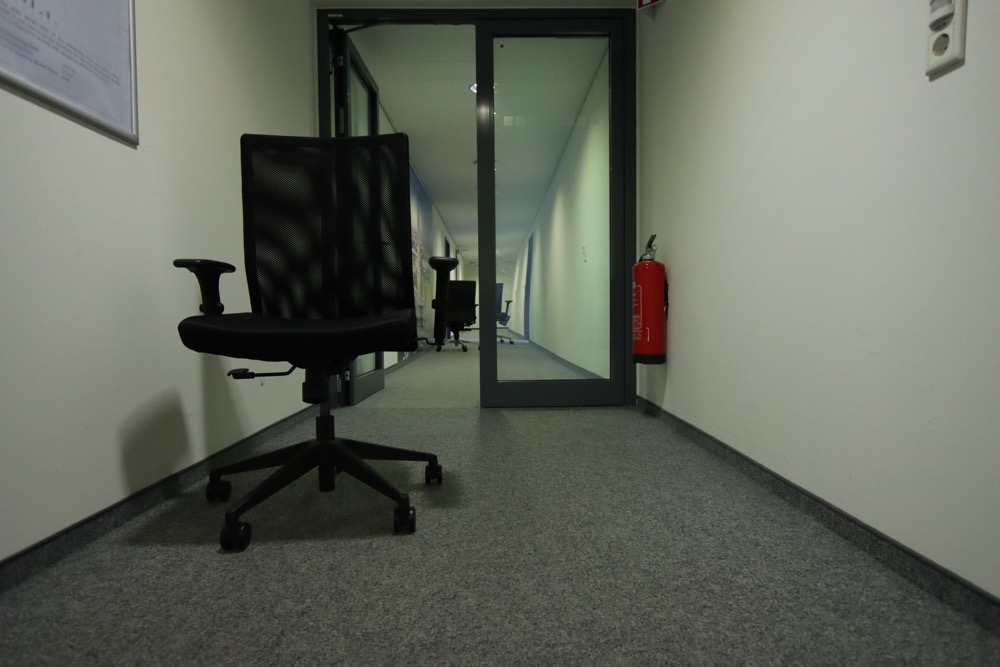}
    & \includegraphics[width=0.24\columnwidth]{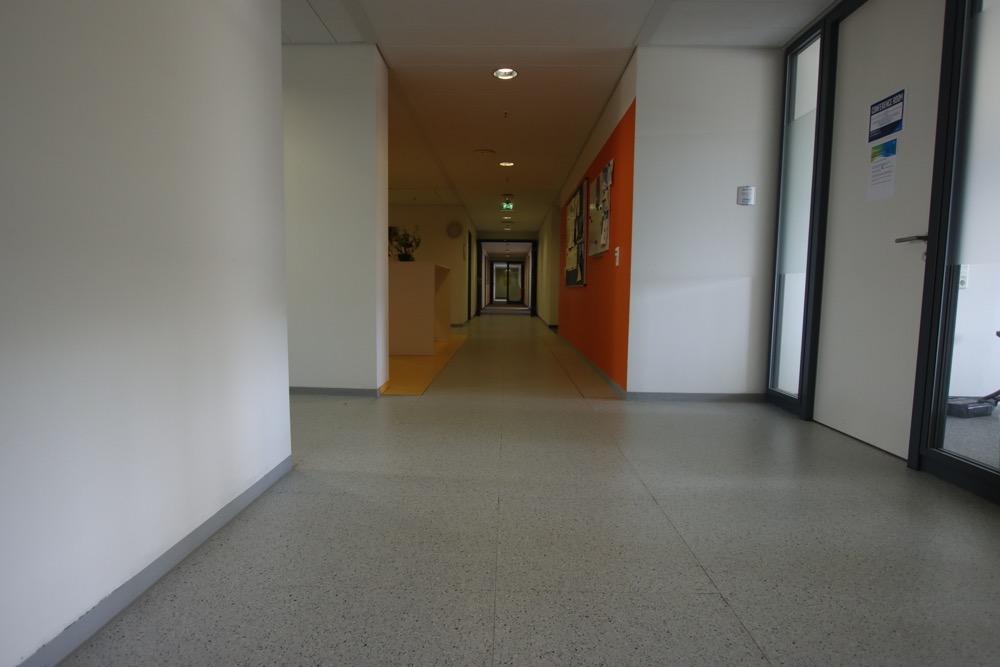}
    & \includegraphics[width=0.24\columnwidth]{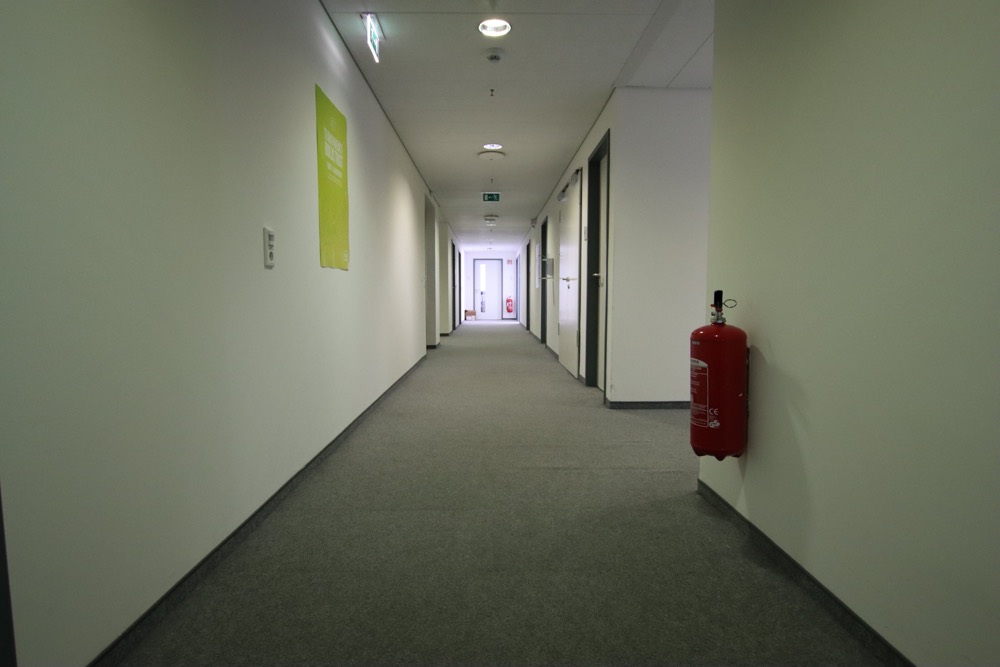}
    & \includegraphics[width=0.24\columnwidth]{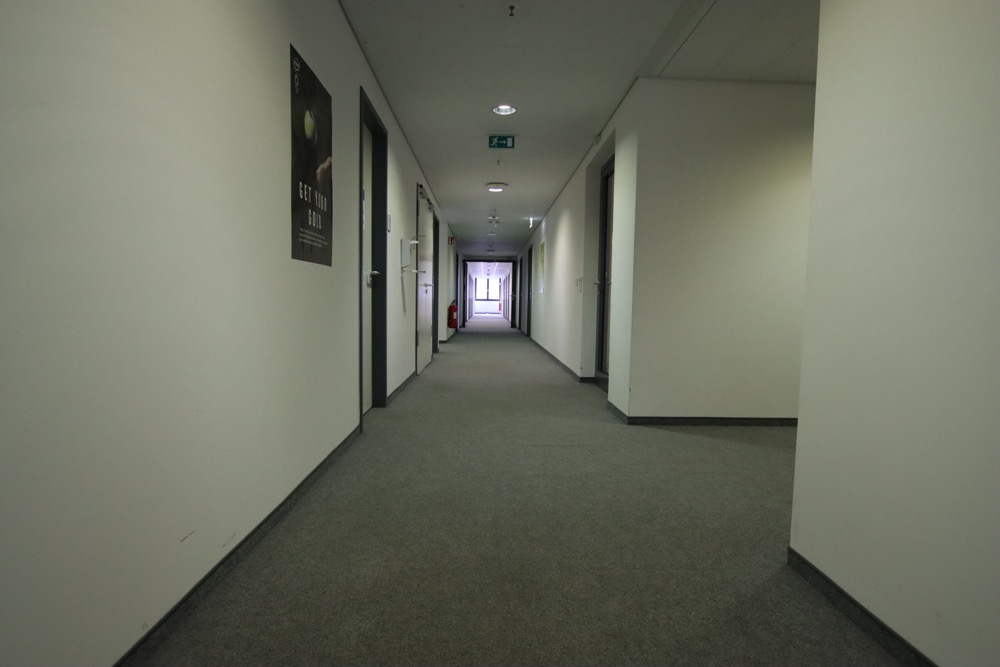} \\
      \includegraphics[width=0.24\columnwidth]{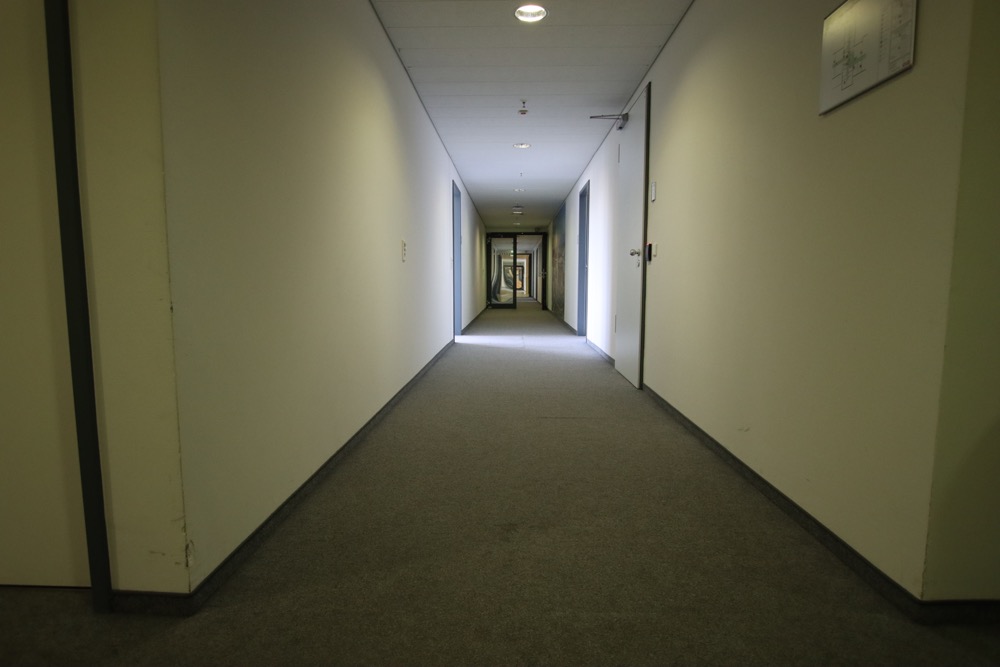}
    & \includegraphics[width=0.24\columnwidth]{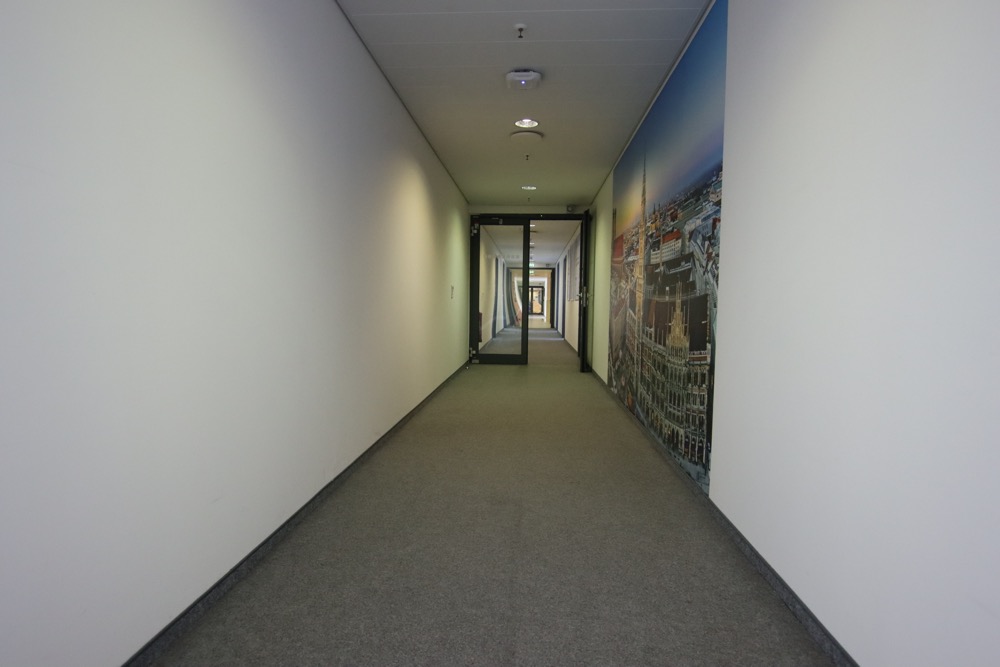}
    & \includegraphics[width=0.24\columnwidth]{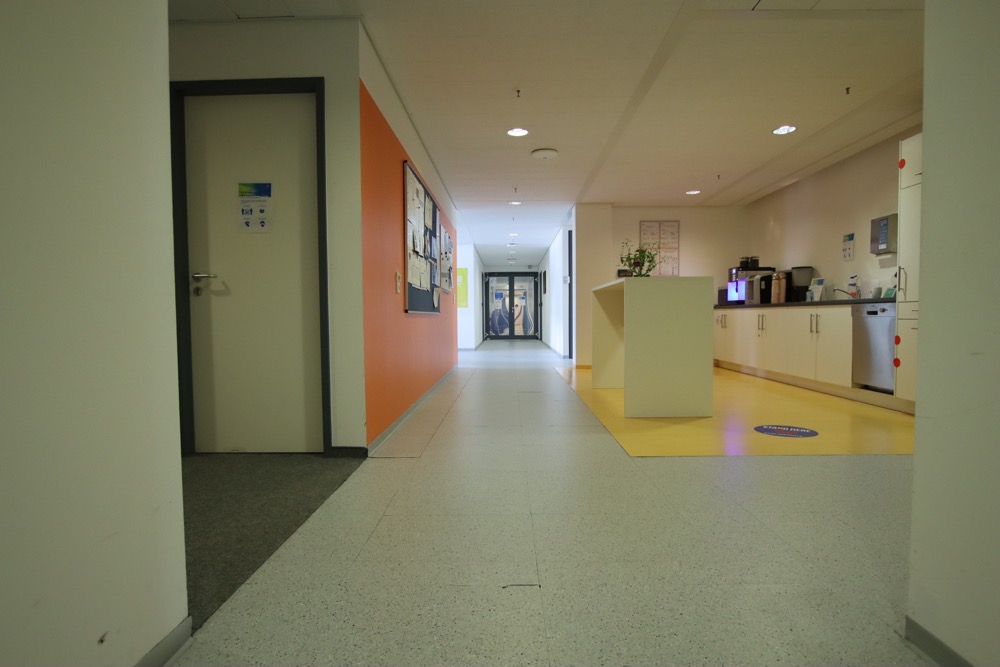}
    & \includegraphics[width=0.24\columnwidth]{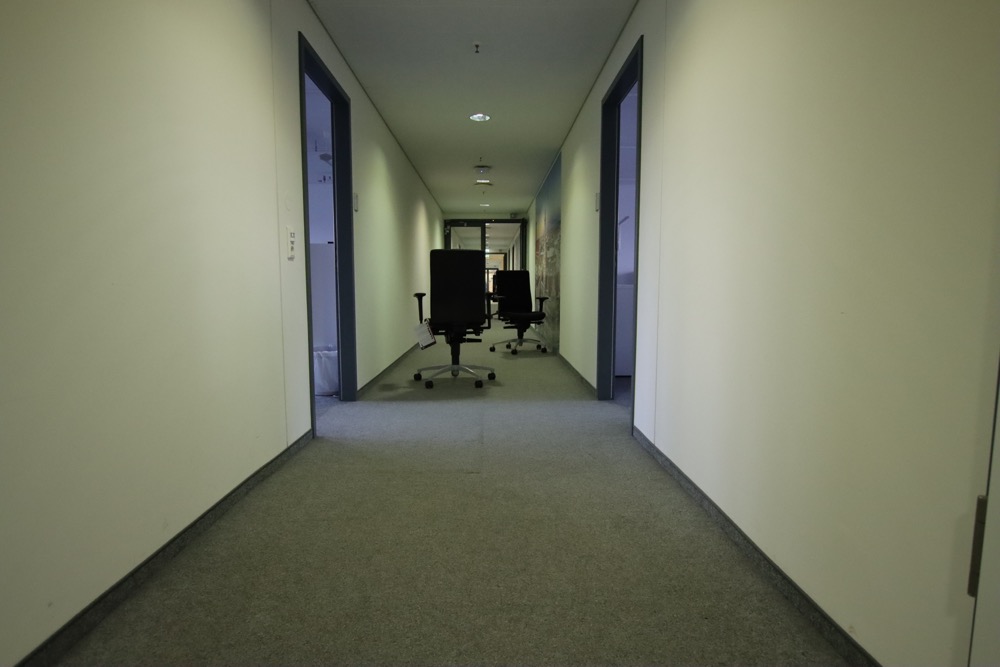} \\
 \end{tabular}
 \caption{\textbf{Training Environment.} The images depict the environment where the training data was collected.}
\label{fig:training_environment}
\end{figure}

We show a map of our training environment in \figLabel \ref{fig:training_routes} and several images in \figLabel \ref{fig:training_environment}. We define three routes and call them \textit{R1}, \textit{R2} and \textit{R3}. \textit{R1} consists of 5 bi-directional segments with a total of $20$ intersections: $8$ left turns, $8$ right turns, and $4$ straights. One data unit corresponds to about $8$ minutes or $12@000$ frames. \textit{R2} and \textit{R3} both consist of 2 bi-directional segments with a total of two left turns, and two right turns at a T-junction. One data unit corresponds to about $70$ seconds or $1@750$ frames.

For the experiments in the paper, we collect a dataset with the Xiaomi Mi9 which consists of two data units from \textit{R1} and six data units from both \textit{R2} and \textit{R3}. Half of the data on \textit{R1} is collected with noise and obstacles and the other without. Half of the data on \textit{R2} and \textit{R3} is collected with noise and the other without. The complete dataset contains approximately $45@000$ frames corresponding to $30$ minutes worth of data.

\subsection{Evaluation Details}
\label{sec:eval_details}
\begin{figure}[!htb]
 \centering
 \includegraphics[trim=0 40 0 90, clip, width=\linewidth]{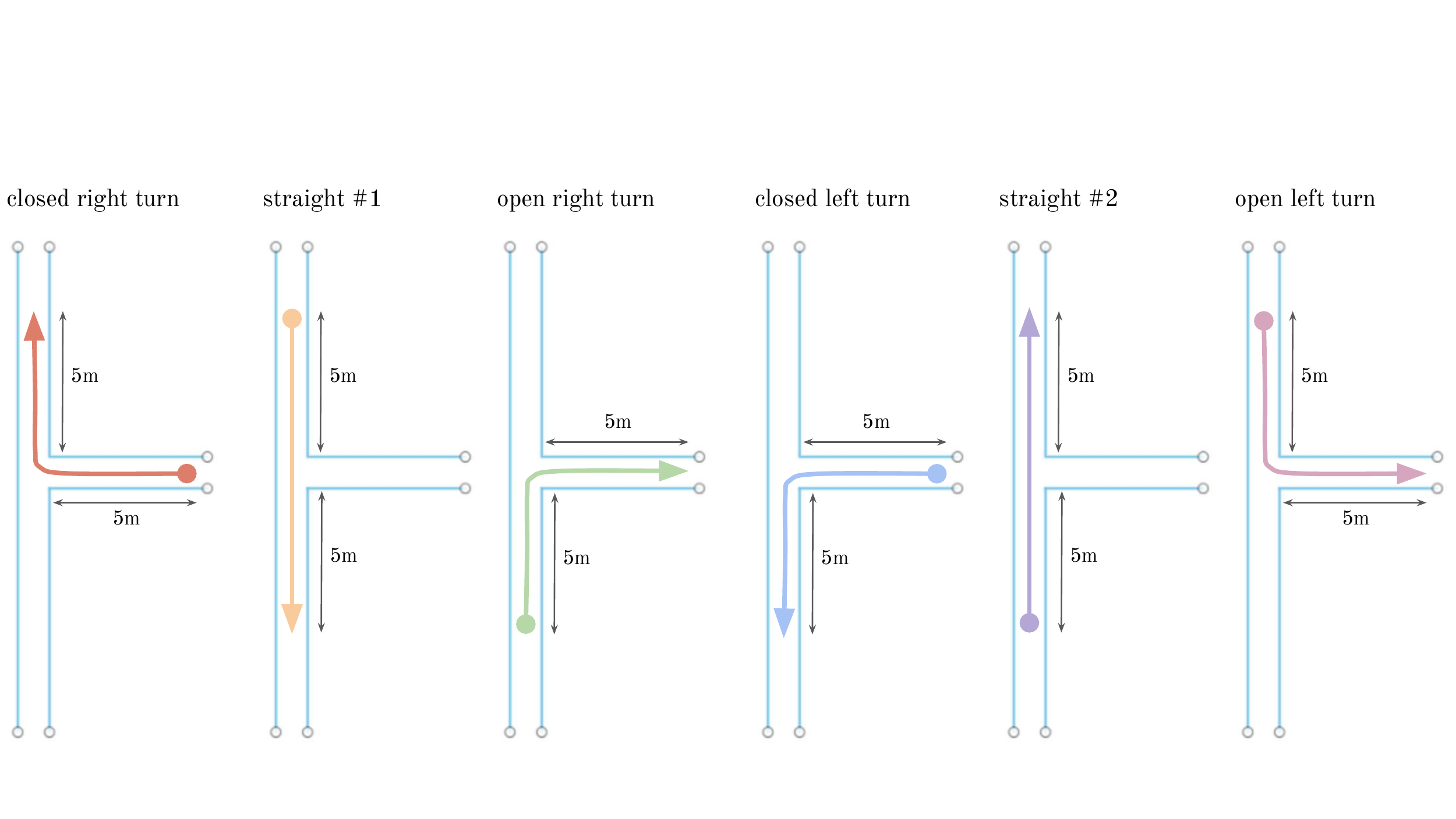}
 \caption{\textbf{Evaluation Route 1: T-junction.} Our evaluation route consists of six segments with a total of two straights, two right turns, and two left turns. We report mean and standard deviation across three trials.}
 \label{fig:evaluation_route_1}
\end{figure}

We design an evaluation setup that is simple to set up in various environments in order to encourage benchmarking using OpenBot. The only thing needed is a T-junction as shown in \figLabel \ref{fig:evaluation_route_1}. We define one trial as six segments comprising two straights, two right turns, and two left turns. We distinguish between closed and open turns, the latter being more difficult. To ensure a simple yet comprehensive comparison, we adopt the following metrics: success, distance, number of collisions and inference speed. Success is a binary value indicating weather or not a segment was completed. The distance is measured along the boundary of a segment without counting the intersections. This way, every segment has a length of $10$m and the metric is invariant to different corridor widths. If an intersection is missed, we measure the distance until the beginning of the intersection (\ie $5$m). The number of collisions is recorded per segment. We measure the inference time of the driving policy per frame to compute the average FPS. All measurements are averaged across the six segments to obtain the results for one trial. We report the mean and standard deviation across three trials for all metrics. All results in the paper are obtained using this evaluation route.

\subsection{Additional Experiments} 
\label{sec:additional_experiments}

For the following experiments, we collect multiple data units along route \textit{R1} in the training environment (\figLabel \ref{fig:training_routes}) with multiple robots and smartphones. We consider a total of four datasets; each dataset consists of $12$ data units or approximately 96 minutes of data, half of which is collected with noise and obstacles. Two datasets are used to investigate the impact of using different phones and the other two to investigate the impact of using different bodies. 

\begin{figure}[!htb]
 \centering
 \includegraphics[trim=20 15 120 70, clip, width=\linewidth]{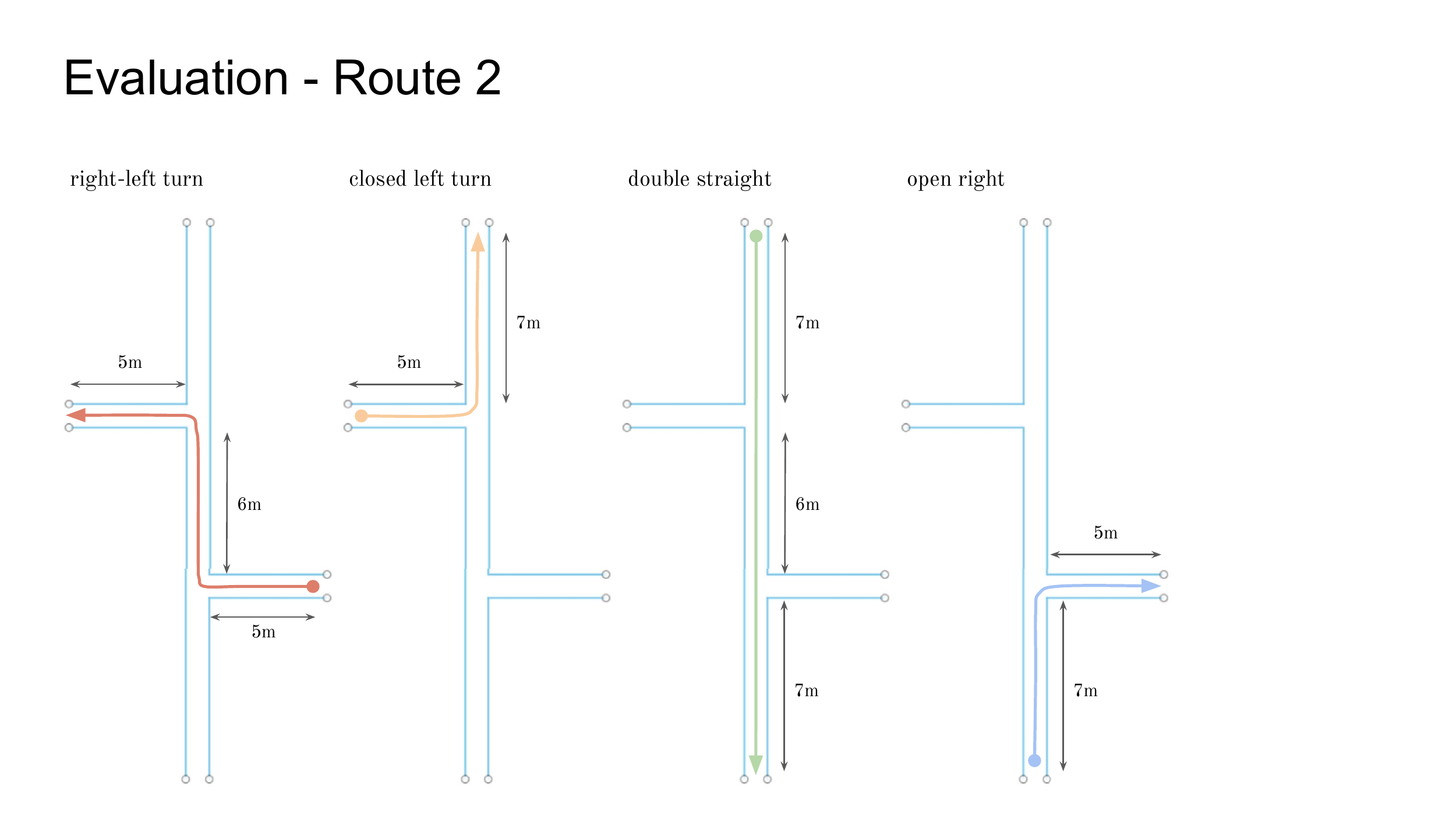}
 \caption{\textbf{Evaluation Route 2: Double T-junction.} Our evaluation route consists of four segments with a total of two straights, two right turns, and two left turns across two intersections. We report mean and standard deviation across three trials.}
 \label{fig:evaluation_route_2}
\end{figure}

Since these policies are trained on more data, we design a more difficult evaluation route as shown in \figLabel \ref{fig:evaluation_route_2}. The route contains the same type of maneuvers, but across two different intersections and divided into less segments. As a result, small errors are more likely to accumulate, leading to unsuccessful segments and a lower average success rate.

\begin{table*}[!htb]
\setlength{\tabcolsep}{7pt}
\centering
\begin{tabular}{l|ccc|ccc|ccc|ccc}
\toprule
Evaluation & \multicolumn{3}{c|}{Mi9} & \multicolumn{3}{c|}{P30 Lite} & \multicolumn{3}{c|}{Pocofone F1} & \multicolumn{3}{c}{Galaxy Note 10}\\
Training & All & Mi9 & $\Delta$ & All & Mi9 & $\Delta$ & All & Mi9 & $\Delta$ & All & Mi9 & $\Delta$\\
\midrule
Distance (\%) $\uparrow$  & 97$\pm$5 & 94$\pm$5 & \textbf{3} & 85$\pm$19 & 80$\pm$6 & \textbf{5} & 79$\pm$7 & 73$\pm$1 & \textbf{6} & 87$\pm$11 & 69$\pm$7 & \textbf{18} \\
Success (\%) $\uparrow$  & 92$\pm$14 & 83$\pm$14 & \textbf{9} & 75$\pm$25 & 50$\pm$0 & \textbf{25} & 42$\pm$14& 42$\pm$14 & \textbf{0} & 67$\pm$14 & 42$\pm$14 & \textbf{25} \\
Collisions $\downarrow$ & 0.0$\pm$0.0 & 0.0$\pm$0.0 & \textbf{0.0} & 1.0$\pm$1.0 & 0.0$\pm$0.0 & \textbf{1.0} & 0.3$\pm$0.6 & 1.3$\pm$0.6 & \textbf{-1.0} & 1.7$\pm$0.6 & 1.3$\pm$0.6 & \textbf{0.4} \\ 
\bottomrule
\end{tabular}
\caption{\textbf{Autonomous navigation: Transfer across smartphones.} We report the mean and standard deviation across three trials. Each trial consists of several segments with a total of 2 straights, 2 left turns, and 2 right turns.}
\label{tbl:phone_experiments}
\vspace{-8pt}
\end{table*}

\subsection{Learning from data collected with multiple smartphones}
We investigate whether training on data from multiple phones helps generalization and robustness. We train two identical driving policies, one on data acquired with six different phones (\tblLabel~\ref{tbl:smartphones}, top) and another with the same amount of data from only one phone, the Xiaomi Mi9; we keep the robot body the same for this set of experiments. We evaluate both policies on the common training phone, the Mi9. We also evaluate both driving policies on three held-out test phones that were not used for data collection and differ in terms of camera sensor and manufacturer (\tblLabel~\ref{tbl:smartphones}, bottom). The P30 Lite has the same camera sensor as the Mi9, but is from a different manufacturer. The Pocofone F1 has a different camera sensor, but is from the same manufacturer. The Galaxy Note 10 differs in both aspects, manufacturer and camera sensor.

The results are summarized in \tblLabel~\ref{tbl:phone_experiments}. 
We find that the driving policy trained on data from multiple phones consistently outperforms the driving policy trained on data from a single phone. This effect becomes more noticeable when deploying the policy on phones from different manufacturers and with different camera sensors. 
However, driving behaviour is sometimes more abrupt which is reflected by the higher number of collisions. This is probably due to the different field-of-views and positions of the camera sensors making learning more difficult. We expect that this will be overcome with more training data.

We also performed some experiments using the low-end Nokia 2.2 phone, which costs about \$100. It is able to run our autonomous navigation network at $10$ frames per second. Qualitatively, the driving performance is similar to the other phones we evaluated. However, since it was unable to make predictions in real time, we did not include it in our main experiments, which were focused on the impact of camera sensor and manufacturer.

\subsection{Learning from data collected with multiple robot bodies}
We also investigate whether training on data from multiple robot bodies helps generalization and robustness. One policy is trained on data collected with three different bodies and another with the same amount of data from a single body; we keep the smartphone fixed for this set of experiments. We evaluate both policies on the common training body, B1, which was used during data collection. We also evaluate on a held-out test body, B4. 

\begin{table}[!htb]
\setlength{\tabcolsep}{4pt}
\centering
\begin{tabular}{l|ccc|ccc}
\toprule
Evaluation & \multicolumn{3}{c|}{Body 1} & \multicolumn{3}{c}{Body 4} \\
Training & B1-B3 & B1 & $\Delta$ & B1-B3 & B1 & $\Delta$\\
\midrule
Distance (\%) $\uparrow$    & 97$\pm$5 & 94$\pm$5 & \textbf{3} & 94$\pm$5 & 92$\pm$8 & \textbf{2} \\
Success (\%) $\uparrow$     & 92$\pm$14 & 83$\pm$14 & \textbf{9} & 83$\pm$14 & 75$\pm$25 & \textbf{8} \\
Collisions $\downarrow$     & 0.0$\pm$0.0 & 0.0$\pm$0.0 & \textbf{0.0} & 0.0$\pm$0.0 & 0.7$\pm$0.6 & \textbf{-0.7}\\ 
\bottomrule
\end{tabular}
\caption{\textbf{Autonomous navigation: Transfer across robot bodies.} We report the mean and standard deviation across three trials. Each trial consists of several segments with a total of 2 straights, 2 left turns, and 2 right turns.}
\label{tbl:body_experiments}
\vspace{-8pt}
\end{table}

The results are summarized in \tblLabel~\ref{tbl:body_experiments}. We find that the driving policy that was trained on multiple robot bodies performs better, especially in terms of success rate, where small mistakes can lead to failure. The policy that was trained on a single body sways from side to side and even collides with the environment when deployed on the test body. The actuation of the bodies is noisy due to the cheap components. Every body responds slightly differently to the control signals. Most bodies have a bias to veer to the left or to the right due to imprecision in the assembly or the low-level controls. The policy trained on multiple bodies learns to be robust to these factors of variability, exhibiting stable learned behavior both on the training bodies and on the held-out test body.

Despite the learned robustness, the control policy is still somewhat vehicle-specific, \eg the differential drive setup and general actuation model of the motors. An alternative would be predicting a desired trajectory instead and using a low-level controller to produce vehicle-specific actions. This can further ease the learning process and lead to more general driving policies. 

\clearpage
\bibliographystyle{IEEEtran}
\bibliography{references}

\begin{thebibliography}{10}
\providecommand{\url}[1]{#1}
\csname url@rmstyle\endcsname
\providecommand{\newblock}{\relax}
\providecommand{\bibinfo}[2]{#2}
\providecommand\BIBentrySTDinterwordspacing{\spaceskip=0pt\relax}
\providecommand\BIBentryALTinterwordstretchfactor{4}
\providecommand\BIBentryALTinterwordspacing{\spaceskip=\fontdimen2\font plus
\BIBentryALTinterwordstretchfactor\fontdimen3\font minus
  \fontdimen4\font\relax}
\providecommand\BIBforeignlanguage[2]{{%
\expandafter\ifx\csname l@#1\endcsname\relax
\typeout{** WARNING: IEEEtran.bst: No hyphenation pattern has been}%
\typeout{** loaded for the language `#1'. Using the pattern for}%
\typeout{** the default language instead.}%
\else
\language=\csname l@#1\endcsname
\fi
#2}}

\bibitem{kau2019stanford}
N.~Kau, A.~Schultz, N.~Ferrante, and P.~Slade, ``Stanford doggo: An
  open-source, quasi-direct-drive quadruped,'' in \emph{ICRA}, 2019.

\bibitem{grimminger2019open}
F.~Grimminger, A.~Meduri, M.~Khadiv, J.~Viereck, M.~W{\"u}thrich, M.~Naveau,
  V.~Berenz, S.~Heim, F.~Widmaier, J.~Fiene, \emph{et~al.}, ``An open
  torque-controlled modular robot architecture for legged locomotion
  research,'' \emph{arXiv:1910.00093}, 2019.

\bibitem{yang2019replab}
B.~Yang, J.~Zhang, V.~Pong, S.~Levine, and D.~Jayaraman, ``Replab: A
  reproducible low-cost arm benchmark platform for robotic learning,''
  \emph{arXiv:1905.07447}, 2019.

\bibitem{gupta2018robot}
A.~Gupta, A.~Murali, D.~P. Gandhi, and L.~Pinto, ``Robot learning in homes:
  Improving generalization and reducing dataset bias,'' in \emph{NeurIPS},
  2018.

\bibitem{gealy2019quasi}
D.~V. Gealy, S.~McKinley, B.~Yi, P.~Wu, P.~R. Downey, G.~Balke, A.~Zhao,
  M.~Guo, R.~Thomasson, A.~Sinclair, \emph{et~al.}, ``Quasi-direct drive for
  low-cost compliant robotic manipulation,'' in \emph{ICRA}, 2019.

\bibitem{balaji2019deepracer}
B.~Balaji, S.~Mallya, S.~Genc, S.~Gupta, L.~Dirac, V.~Khare, G.~Roy, T.~Sun,
  Y.~Tao, B.~Townsend, \emph{et~al.}, ``Deepracer: Educational autonomous
  racing platform for experimentation with sim2real reinforcement learning,''
  \emph{arXiv:1911.01562}, 2019.

\bibitem{robomaster_s1}
{DJI Robomaster S1}, ``\url{https://www.dji.com/robomaster-s1},'' accessed:
  2020-06-20.

\bibitem{nvidia_jetbot}
{Nvidia JetBot}, ``\url{https://github.com/nvidia-ai-iot/jetbot},'' accessed:
  2020-06-20.

\bibitem{paull2017duckietown}
L.~Paull, J.~Tani, H.~Ahn, J.~Alonso-Mora, L.~Carlone, M.~Cap, Y.~F. Chen,
  C.~Choi, J.~Dusek, Y.~Fang, \emph{et~al.}, ``Duckietown: an open, inexpensive
  and flexible platform for autonomy education and research,'' in \emph{ICRA},
  2017.

\bibitem{ignatov2019ai}
A.~Ignatov, R.~Timofte, A.~Kulik, S.~Yang, K.~Wang, F.~Baum, M.~Wu, L.~Xu, and
  L.~Van~Gool, ``{AI} benchmark: All about deep learning on smartphones in
  2019,'' in \emph{ICCV Workshops}, 2019.

\bibitem{rubenstein2012kilobot}
M.~Rubenstein, C.~Ahler, and R.~Nagpal, ``Kilobot: A low cost scalable robot
  system for collective behaviors,'' in \emph{ICRA}, 2012.

\bibitem{mclurkin2014robot}
J.~McLurkin, A.~McMullen, N.~Robbins, G.~Habibi, A.~Becker, A.~Chou, H.~Li,
  M.~John, N.~Okeke, J.~Rykowski, \emph{et~al.}, ``A robot system design for
  low-cost multi-robot manipulation,'' in \emph{IROS}, 2014.

\bibitem{wilson2016pheeno}
S.~Wilson, R.~Gameros, M.~Sheely, M.~Lin, K.~Dover, R.~Gevorkyan, M.~Haberland,
  A.~Bertozzi, and S.~Berman, ``Pheeno, a versatile swarm robotic research and
  education platform,'' \emph{IEEE Robotics and Automation Letters}, vol.~1,
  no.~2, pp. 884--891, 2016.

\bibitem{gonzales2016autonomous}
J.~Gonzales, F.~Zhang, K.~Li, and F.~Borrelli, ``Autonomous drifting with
  onboard sensors,'' in \emph{Advanced Vehicle Control}, 2016.

\bibitem{codevilla2018end}
F.~Codevilla, M.~M{\"u}ller, A.~L{\'o}pez, V.~Koltun, and A.~Dosovitskiy,
  ``End-to-end driving via conditional imitation learning,'' in \emph{ICRA},
  2018.

\bibitem{srinivasa2019mushr}
S.~S. Srinivasa, P.~Lancaster, J.~Michalove, M.~Schmittle, C.~S.~M. Rockett,
  J.~R. Smith, S.~Choudhury, C.~Mavrogiannis, and F.~Sadeghi, ``Mushr: A
  low-cost, open-source robotic racecar for education and research,''
  \emph{arXiv:1908.08031}, 2019.

\bibitem{o2019f1}
M.~O'Kelly, V.~Sukhil, H.~Abbas, J.~Harkins, C.~Kao, Y.~V. Pant, R.~Mangharam,
  D.~Agarwal, M.~Behl, P.~Burgio, \emph{et~al.}, ``F1/10: An open-source
  autonomous cyber-physical platform,'' \emph{arXiv:1901.08567}, 2019.

\bibitem{goldfain2019autorally}
B.~Goldfain, P.~Drews, C.~You, M.~Barulic, O.~Velev, P.~Tsiotras, and J.~M.
  Rehg, ``Autorally: An open platform for aggressive autonomous driving,''
  \emph{IEEE Control Systems Magazine}, vol.~39, pp. 26--55, 2019.

\bibitem{riedo2013thymio}
F.~Riedo, M.~Chevalier, S.~Magnenat, and F.~Mondada, ``Thymio ii, a robot that
  grows wiser with children,'' in \emph{IEEE Workshop on Advanced Robotics and
  its Social Impacts}, 2013.

\bibitem{lee2019device}
J.~Lee, N.~Chirkov, E.~Ignasheva, Y.~Pisarchyk, M.~Shieh, F.~Riccardi,
  R.~Sarokin, A.~Kulik, and M.~Grundmann, ``On-device neural net inference with
  mobile {GPUs},'' \emph{arXiv:1907.01989}, 2019.

\bibitem{phonebot15}
S.~Owais, ``Turn your phone into a robot,''
  \url{https://www.instructables.com/id/Turn-Your-Phone-into-a-Robot/}, 2015,
  accessed: 2020-06-20.

\bibitem{androidRC16}
M.~Rovai, ``Hacking a rc car to control it using an android device,''
  \url{https://www.hackster.io/mjrobot/hacking-a-rc-car-to-control-it-using-an-android-device-7d5b9a},
  2016, accessed: 2020-06-20.

\bibitem{botiful12}
C.~Delaunay, ``Botiful, social telepresence robot for android,''
  \url{https://www.kickstarter.com/projects/1452620607/botiful-telepresence-robot-for-android},
  2012, accessed: 2020-06-20.

\bibitem{romo12}
Romotive, ``Romo - the smartphone robot for everyone,''
  \url{https://www.kickstarter.com/projects/peterseid/romo-the-smartphone-robot-for-everyone},
  2012, accessed: 2020-06-20.

\bibitem{ethos15}
xCraft, ``Phonedrone ethos - a whole new dimension for your smartphone,''
  \url{https://www.kickstarter.com/projects/137596013/phonedrone-ethos-a-whole-new-dimension-for-your-sm},
  2015, accessed: 2020-06-20.

\bibitem{wheelphone13}
GCtronic, ``Wheelphone,'' \url{http://www.wheelphone.com}, 2013, accessed:
  2020-06-20.

\bibitem{yim2010}
J.~{Yim}, S.~{Chun}, K.~{Jung}, and C.~D. {Shaw}, ``Development of
  communication model for social robots based on mobile service,'' in
  \emph{International Conference on Social Computing}, 2010.

\bibitem{setapen2012creating}
A.~Setapen, ``Creating robotic characters for long-term interaction,'' Ph.D.
  dissertation, MIT, 2012.

\bibitem{cao2019v}
Y.~Cao, Z.~Xu, F.~Li, W.~Zhong, K.~Huo, and K.~Ramani, ``V. ra: An in-situ
  visual authoring system for robot-iot task planning with augmented reality,''
  in \emph{DIS}, 2019.

\bibitem{oros2013smartphone}
N.~Oros and J.~L. Krichmar, ``Smartphone based robotics: Powerful, flexible and
  inexpensive robots for hobbyists, educators, students and researchers,''
  \emph{IEEE Robotics \& Automation Magazine}, vol.~1, p.~3, 2013.

\bibitem{tf_od_app}
{Tensorflow Object Detection Android Application},
  ``\url{https://github.com/tensorflow/examples/tree/master/lite/examples/object_detection/android},''
  accessed: 2020-06-20.

\bibitem{karaman2017project}
S.~Karaman, A.~Anders, M.~Boulet, J.~Connor, K.~Gregson, W.~Guerra, O.~Guldner,
  M.~Mohamoud, B.~Plancher, R.~Shin, \emph{et~al.}, ``Project-based,
  collaborative, algorithmic robotics for high school students: Programming
  self-driving race cars at mit,'' in \emph{ISEC}, 2017.

\bibitem{donkey_car}
W.~Roscoe, ``An opensource diy self driving platform for small scale cars,''
  \url{https://www.donkeycar.com}, accessed: 2020-06-20.

\bibitem{dekan2013irobot}
M.~Dekan, F.~Ducho{\v{n}}, L.~Juri{\v{s}}ica, A.~Vitko, and A.~Babinec,
  ``irobot create used in education,'' \emph{Journal of Mechanics Engineering
  and Automation}, vol.~3, no.~4, pp. 197--202, 2013.

\bibitem{rubenstein2015aerobot}
M.~Rubenstein, B.~Cimino, R.~Nagpal, and J.~Werfel, ``Aerobot: An affordable
  one-robot-per-student system for early robotics education,'' in \emph{ICRA},
  2015.

\bibitem{bojarski2016end}
M.~Bojarski, D.~Del~Testa, D.~Dworakowski, B.~Firner, B.~Flepp, P.~Goyal, L.~D.
  Jackel, M.~Monfort, U.~Muller, J.~Zhang, \emph{et~al.}, ``End to end learning
  for self-driving cars,'' \emph{arXiv:1604.07316}, 2016.

\bibitem{howard2017mobilenets}
A.~G. Howard, M.~Zhu, B.~Chen, D.~Kalenichenko, W.~Wang, T.~Weyand,
  M.~Andreetto, and H.~Adam, ``Mobilenets: Efficient convolutional neural
  networks for mobile vision applications,'' \emph{arXiv:1704.04861}, 2017.

\bibitem{howard2019searching}
A.~Howard, M.~Sandler, G.~Chu, L.-C. Chen, B.~Chen, M.~Tan, W.~Wang, Y.~Zhu,
  R.~Pang, V.~Vasudevan, \emph{et~al.}, ``Searching for mobilenetv3,'' in
  \emph{ICCV}, 2019.

\bibitem{lin2014microsoft}
T.-Y. Lin, M.~Maire, S.~Belongie, J.~Hays, P.~Perona, D.~Ramanan,
  P.~Doll{\'a}r, and C.~L. Zitnick, ``Microsoft coco: Common objects in
  context,'' in \emph{ECCV}, 2014.

\bibitem{codevilla2018offline}
F.~Codevilla, A.~M. Lopez, V.~Koltun, and A.~Dosovitskiy, ``On offline
  evaluation of vision-based driving models,'' in \emph{ECCV}, 2018.

\end{thebibliography}

\end{document}